\documentclass[journal]{IEEEtran}
\usepackage{amsmath}
\usepackage{amsfonts} 
\usepackage{bm}
\usepackage{amsthm}
\usepackage{amsmath}
\usepackage{amssymb}
\usepackage{graphicx}
\usepackage{subfigure}
\usepackage{multirow}
\usepackage{setspace}
\usepackage{amsfonts}
\usepackage{cite}
\usepackage{bm}
\usepackage{booktabs}
\usepackage{makecell}

\newtheorem*{theorem*}{Theorem}
\newtheorem*{lemma*}{Lemma}
\newtheorem*{assumption*}{Assumption}

\begin{document}
\title{A Robust 3D Registration Method via Simultaneous Inlier Identification and Model Estimation}
\author{Xianyun Qian, Fei Wen, and Peilin Liu
\IEEEcompsocitemizethanks{
\IEEEcompsocthanksitem X. Qian, F. Wen  and P. Liu are with the 
School of Information Science and Electronic Engineering / School of Integrated Circuits, Shanghai Jiao Tong University, Shanghai, China. 
E-mail: \{qianxianyun, wenfei, liupeilin\}@sjtu.edu.cn.
}
\thanks{}}
        
\markboth{}
{Shell \MakeLowercase{\textit{et al.}}: Bare Demo of IEEEtran.cls for Journals}

\maketitle

\begin{abstract}
Robust 3D registration is a fundamental problem in computer vision and robotics, where the goal is to estimate the geometric transformation between two sets of measurements in the presence of noise, mismatches, and extreme outlier contamination. Existing robust registration methods are mainly built on either maximum consensus (MC) estimators, which first identify inliers and then estimate the transformation, or M-estimators, which directly optimize a robust objective. In this work, we revisit a truncated-loss based formulation for simultaneous inlier identification and model estimation (SIME) and study it in the context of 3D registration. We show that, compared with MC-based robust fitting, SIME can achieve a lower fitting residual because it incorporates residual magnitudes into the inlier selection process. To solve the resulting nonconvex problem, we develop an alternating minimization (AM) algorithm, and further propose an AM method embedded with semidefinite relaxation (SDR) to alleviate the difficulty caused by the binary inlier variables. We instantiate the proposed framework for 3D rotation search and rigid point-set registration using quaternion-based formulations. Experimental results on both simulated and real-world registration tasks demonstrate that the proposed methods compare favorably with strong baseline solvers, especially in challenging cases with high noise levels and many outliers.
\end{abstract}

\section{Introduction}
\label{sec:introduction}

Robust 3D registration is a central problem in computer vision, robotics, and geometric perception \cite{3D-01,3D-02,3D-03}. Given a set of tentative correspondences between two 3D data sets, the goal is to estimate the underlying geometric transformation, such as a 3D rotation or a full rigid motion, despite the presence of measurement noise and outlier correspondences. Robust 3D registration is a key component in a wide range of applications, including point cloud alignment, object pose estimation, robot localization, and multi-view reconstruction.

A major challenge in 3D registration lies in the large number of outliers that often arise in practical correspondence generation. For example, feature matching between point clouds or scans can easily produce incorrect correspondences due to repetitive structures, partial overlap, occlusion, or sensor noise. As a result, standard least-squares estimation is highly sensitive to mismatches, and robust estimators are required.

Among robust approaches, maximum consensus (MC) methods and M-estimators are two of the most widely used frameworks. MC-based methods aim to find a transformation that is consistent with as many correspondences as possible under a prescribed inlier threshold. Classic RANSAC \cite{05} and its variants \cite{06,07,08,09,10,57,58} are representative randomized MC methods and remain popular due to their simplicity and broad applicability. For 3D registration, they are often used together with geometric verification to reject false matches. However, their solution quality depends on random sampling and the iteration budget, which makes their performance unpredictable, especially when the outlier ratio is high. Global MC methods based on branch-and-bound, tree search, or enumeration \cite{11,12,13,14,15,16} can provide optimality guarantees, but their computational cost limits scalability. Deterministic approximate MC methods, such as convex relaxation, reweighted $\ell_1$, $\ell_0$, and exact penalty methods \cite{17,18,19,20,21}, offer a more efficient alternative, but their performance can still degrade in highly contaminated settings.

Another important line of work is M-estimation, where the transformation is estimated by directly minimizing a robust loss function \cite{26,27,28,29,30}. Compared with MC, M-estimators avoid explicit combinatorial inlier-set search and often lead to efficient iterative solvers. However, most M-estimators mainly focus on robust parameter estimation rather than explicit inlier/outlier separation, and their effectiveness depends strongly on the choice of loss and optimization strategy.

A particularly attractive class of M-estimators is based on truncated loss. Truncated least-squares and related formulations possess two desirable properties: strong robustness against outliers, and enabling inliers/outliers identification. These properties have led to several influential robust registration methods in recent years. For example, semidefinite relaxation has been used for robust rotation estimation under extreme outlier rates \cite{31}, graduated nonconvexity has been adopted to improve scalability \cite{29}, and  certifiable schemes have been proposed for rigid point cloud registration \cite{51}. Despite this progress, there is still a need for a unified and efficient truncated-loss framework that can directly couple inlier identification with model estimation and can be applied to both rotation-only and full Euclidean registration.

In this paper, we investigate a truncated-loss formulation for \emph{simultaneous inlier identification and model estimation} (SIME) for robust 3D registration. It jointly determines which correspondences should be treated as inliers and what transformation best explains them. This is particularly appealing for 3D registration, because the quality of the estimated transformation and the correctness of the inlier set are strongly coupled.

First, we revisit the SIME formulation in a general form and then specialize it to 3D registration problems with quaternion-based rotation representation. In contrast to MC estimators, SIME also takes the residual magnitude into account when determining the inlier set. We show that the lowest achievable fitting residual of SIME is lower than that of MC-based robust fitting. From the perspective of registration accuracy, this makes SIME especially attractive in difficult scenarios with high noise and many outliers.

Second, we propose two algorithms for the SIME formulation, one is based on the direct alternating minimization (AM) algorithm, and the other additionally uses a semidefinite relaxation (SDR) of the binary variables to ease the high non-convexity of the problem. Further, the sparsity of the problem is exploited in combination with low-rank factorization to achieve high efficiency. 

Third, we evaluate the proposed methods on both rotation registration and 6-DoF rigid registration tasks. Experiments on simulated and real-world data show that, compared with strong baseline methods, the proposed algorithms achieves favorable strong performance particularly in challenging settings with high noise and a large fraction of incorrect correspondences.

\textit{Notations:} $\mathbb{I}(\cdot)   \in  {\{0,1\}}$ is the indicator function.
$ \otimes $ and $\circ$ stand for the Kronecker product and quaternion product, respectively.
${{\bf{1}}_N}$  is an $N$ dimension vector with all elements being unit,
and ${\bf{0}}$ is a zero vector or matrix with a proper size.
$\left\|  \cdot  \right\|$  and ${\left\|  \cdot  \right\|_p}$ denotes the Euclidean and $\ell_p$ norm, respectively.
$(\cdot)^T$ denotes transpose.
${\mathbb{S}^N}$ is the set of $N\times N$ real-valued symmetric matrices.
For ${\bf{X}}\in{\mathbb{S}^N}$, ${\bf{X}}\succeq{\bf{0}}$ means that ${\bf{X}}$ is semidefinite.
Both ${\bf{X}}(i,j)$ and ${{X}}_{i,j}$ denote the $(i,j)$-th element of the matrix ${\bf{X}}$.
${\rm{vec}}( \cdot )$ is the vectorization operation of a matrix.

\section{Robust Fitting Using Truncated Loss}

Traditionally, maximum consensus (MC) methods are widely used for 3D registration to first identify inliers and then fit a model on the inliers. Here we consider simultaneous inlier identification and model
estimation (SIME) using truncated loss. We first compare these two criteria and then instantiate SIME for 3D registration.

\subsection{Simultaneous Inlier Identification and Model
Estimation Using Truncated Loss}
Truncated loss based M-estimation can be traced back to the works \cite{59,60,61},
where truncated LS loss is used to construct robust estimators with influence functions going to zero.
Recently, truncated LS loss has been used to develop robust algorithms for rotation searching \cite{31} and point cloud registration \cite{51}, which can tolerate extreme amounts of outliers.

Given a set of $N$ measurements, let $\Phi$ denote a generalized loss function to be truncated and $\beta$ denote the truncating level. Using such a truncated loss to estimate a mode parameterized by ${\boldsymbol{\theta}}  \in {\mathbb{R}^d}$, the SIME formulation considered in this work is
\begin{equation} \label{eq_sime}
\begin{split}
&\mathop {{\text{min }}}\limits_{{\boldsymbol{\theta}} \in {\mathbb{R}^d},{\mathbf{s}} \in {{\{ 0,1\} }^N}} \sum\nolimits_{i = 1}^N {(1 \!-\! {s_i})\Phi ({r_i}({\boldsymbol{\theta}})) + \beta {s_i}}\\
&~~~~~~~~{\rm{s.t.}}~~~~~h({\boldsymbol{\theta}})=0,
\end{split}
\end{equation}
where ${r_i}({\boldsymbol{\theta}})$ gives the non-negative residual of the $i$-th measurement with respect to ${\boldsymbol{\theta}}$.
$h({\boldsymbol{\theta}})=0$ is a constraint on the model parameter. For example, as will be detailed in Section 4,
a constraint $\|{\boldsymbol{\theta}}\|=1$ is present in rotation registration and Euclidean registration problems,
in which case $h({\boldsymbol{\theta}})=\|{\boldsymbol{\theta}}\|-1$. The objective in (\ref{eq_sime}) is a truncated cost as $\mathop {\min }_{{s_i} \in \{ 0,1\} } (1\!-\!{s_i})\Phi ( \cdot ) + \beta {s_i}$ is equivalent to ${\text{min}}\left( {\Phi ( \cdot ),\beta } \right)$ \cite{31}. When $\Phi$ is the LS loss, problem (\ref{eq_sime}) without the constraint $h({\boldsymbol{\theta}})=0$ becomes the minimum truncated LS formulation \cite{60}.

Note that we consider a generalized truncated loss rather than the truncated LS loss. While the LS loss is optimal for Gaussian noise, a generalized $\Phi$ also adapts to $\ell_p$ loss with $p<2$, which is desirable when the inlier noise is not Gaussian but super-Gaussian (detailed in Appendix \ref{Lp loss}).

A feature of SIME is it can identify inliers/outliers based on the solution $\bf{s}$,
which is similar to the MC criterion.
Given a set of $N$ measurements, MC aims to find a feasible model parameterized by ${\boldsymbol{\theta}}$,
that is consistent with as many of the measurements as possible up to an inlier residual threshold $\tau > 0$,
i.e., has the largest consensus set $I$ as \cite{01}
\begin{equation}\label{eq_mc}
\begin{split}
& ~~~~~~~~\mathop {{\text{max }}}\limits_{{\boldsymbol{\theta}} \in {\mathbb{R}^d},I \subseteq \Omega } |I|\\
{\rm{s.t.}} ~~&{r_i}({\boldsymbol{\theta}}) \leq \tau, ~~\forall i \in I,~~h({\boldsymbol{\theta}})=0,
\end{split}
\end{equation}
where $\Omega  = \{ 1,2, \cdots ,N\}$ is the index set of the measurements.
Let ${I^*}$ be a solution of (\ref{eq_mc}), then ${I^*}$ stands for the index set of the true inliers,
while the complementary set of ${I^*}$, denoted by $\Omega \backslash {I^*}$, stands for the index set of the true outliers.
Using some auxiliary binary variables ${\mathbf{s}} = {[{s_1}, \cdots ,{s_N}]^T} \in {\{ 0,1\} ^N}$, (\ref{eq_mc}) can be reformulated into
\begin{equation}\label{eq_mc_2}
\begin{split}
& ~~~~~~~~\mathop {{\text{min }}}\limits_{{\boldsymbol{\theta}} \in {\mathbb{R}^d},{\mathbf{s}} \in {{\{ 0,1\} }^N}} \sum\nolimits_{i = 1}^N {{s_i}}\\
{\rm{s.t.}} &~~{r_i}({\boldsymbol{\theta}}) \leq \tau  + {s_i}L,~~\forall i \in \Omega,~~h({\boldsymbol{\theta}})=0,
\end{split}
\end{equation}
where $L$ is a sufficiently large positive constant.

The similarity and difference between SIME and MC can be observed from (\ref{eq_sime}) and (\ref{eq_mc_2}).
Similar to SIME, for a solution of MC, the $i$-th measurement is identified as an outlier if $s_i=1$, or an inlier if $s_i=0$.
While MC does not take the inlier residual into account in inlier identification,
SIME takes that into consideration via simultaneous inlier identification and model fitting in a single step.
Besides, unlike that the ${\boldsymbol{\theta}}$ solution of MC is only a feasible solution satisfying the residual constraint,
the ${\boldsymbol{\theta}}$ solution of SIME is the desired model fitting result on the identified inliers.

SIME uses $\beta$ to balance the number of outliers against the fitting residual,
by which the residual constraint in MC is removed. Obviously,
a larger $\beta$ would result in a larger number of $s_i$'s being zero, i.e., more identified inliers (which are not necessary to be true inliers).
That is a too large $\beta$ would yield a larger consensus size than $I^{*}$ and include some true outliers, while a too small $\beta$ would yield a smaller consensus size than $I^*$  and exclude some true inliers. As SIME estimates the model only using the measurements with residuals less than $\beta$,
$\beta$ can be easily selected as a threshold which tightly upper bounds the fitting residual of the true inliers.
For example, if there exists a proper bound $\tau$ such that $r_i({\boldsymbol{\theta}})\leq\tau$ for $\forall i\in I^*$
with a high probability for the true model, it is reasonable to choose $\beta=\Phi(\tau)$.

The following result shows SIME can be
viewed as an approximate maximum likelihood (ML) estimator
under certain distribution assumption of inliers and outliers.

\textit{\textbf{Proposition 1.}
Denote ${r_i} = {r_i}({\boldsymbol{\theta}})$ for succinctness. SIME admits an approximate maximum-likelihood interpretation under the following conditions:
{i)} For any true inlier, i.e. $\forall i \in I^*$, the residual density has the form
\[
p_{\rm in}(r_i\mid {\boldsymbol{\theta}})=c_i \exp\!\left(-\Phi(r_i,\sigma_i)\right),\qquad r_i\ge 0,
\]
where $c_i>0$ is a normalizing constant, $\sigma_i>0$ is a scale parameter, and $\Phi(\cdot,\sigma_i)$ is nondecreasing on $[0,+\infty)$.
The model parameter is constrained by $h({\boldsymbol{\theta}})=0$.
{ii)} For any true outlier, i.e. $\forall i \in \Omega\backslash I^*$, the residual ${r_i}$ is approximately uniformly distributed on $[\tau,u]$, where $u>\tau$ is an upper bound on the outlier perturbation (the possible maximal perturbation of outliers).
}

Proof of Proposition 1 is given in Appendix \ref{proof:propo1}.

\textit{\textbf{Remark 1.}
This proposition suggests that SIME admits an approximate maximum-likelihood interpretation when the inlier residuals are relatively small and their density can be written in the form $p_{\rm in}(r_i)\propto \exp(-\Phi(r_i,\sigma_i))$, while the outlier residuals are approximately uniformly distributed over a bounded interval. This model covers a broad class of residual distributions. For example, when the inlier residual model is induced by Gaussian noise, $\Phi$ reduces to a quadratic form $\Phi ({r_i},{\sigma _i}) = \frac{{r_i^2}}{{2\sigma _i^2}} + \frac{1}{2}\log (2\pi \sigma _i^2)$, corresponding to the least-squares loss up to additive constants. It also accommodates super-Gaussian noise, for which $\Phi(r_i)\propto r_i^p$ with $p<2$, indicating that an $\ell_p$ loss is more appropriate in the case the inlier noise is super-Gaussian.}

\subsection{Comparison between MC and SIME}

This subsection compares MC with SIME to show that they can produce the same inlier set in special conditions,
but not in general conditions. Particularly, SIME has lower fitting residual than MC fitting.

Before proceeding to the results, we present some definitions will be used in the analysis. Denote
\[{{\boldsymbol{\theta}}_I}: = {\text{arg}}\mathop {{\text{min }}}\limits_{{\boldsymbol{\theta}} \in {\mathbb{R}^d},h({\boldsymbol{\theta}})=0} \sum\limits_{i \in I} {\Phi ({r_i}({\boldsymbol{\theta}}))},\]
\[R(I): = \sum\limits_{i \in I} {\Phi ({r_i}({{\boldsymbol{\theta}}_I}))}=\mathop {{\text{min }}}\limits_{{\boldsymbol{\theta}} \in {\mathbb{R}^d},h({\boldsymbol{\theta}})=0} \sum\limits_{i \in I} {\Phi ({r_i}({\boldsymbol{\theta}}))},\]
and the objective of SIME as
\[f({\boldsymbol{\theta}},{\mathbf{s}}): = \sum\nolimits_{i = 1}^N {(1 - {s_i})\Phi ({r_i}({\boldsymbol{\theta}})) + \Phi (\tau ){s_i}}. \]

As the objective of SIME is separable,
for any fixed ${\boldsymbol{\theta}}$, the optimal choice of each binary variable is given by
\[
s_i = \arg\min_{s\in\{0,1\}} (1-s)\Phi(r_i({\boldsymbol{\theta}}))+\Phi(\tau)s.
\]
Hence, for any global solution $({\boldsymbol{\theta}}^\star,{\mathbf{s}}^\star)$ of SIME with inlier set
\[
 I^\star:=\{i: s_i^\star=0\},
\]
it necessarily holds that
\[
\Phi(r_i({\boldsymbol{\theta}}^\star))\le \Phi(\tau),\quad \forall i\in  I^\star,
\]
and
\[
\Phi(r_i({\boldsymbol{\theta}}^\star))\ge \Phi(\tau),\quad \forall i\in \Omega\backslash I^\star.
\]
Accordingly, we define the collection of self-consistent inlier sets associated with SIME as
\[
\begin{gathered}
\digamma:=\{I\subseteq\Omega:\ \Phi(r_i({\boldsymbol{\theta}}_I))\le \Phi(\tau),\ \forall i\in I;\\
\hspace{5.8em}\Phi(r_i({\boldsymbol{\theta}}_I))\ge \Phi(\tau),\ \forall i\in \Omega\backslash I\}.
\end{gathered}
\]
For any $I\subseteq \Omega$, define
\[
{\mathbf{s}}_I:=[\mathbb{I}(1\notin I),\mathbb{I}(2\notin I),\cdots,\mathbb{I}(N\notin I)]^T.
\]
Then, $({{\boldsymbol{\theta}}_I},{{\mathbf{s}}_I})$ is a 
candidate support set that can correspond to a global SIME optimum only when $I \in \digamma$.
That is $I\in\digamma$ is a necessary self-consistency condition for $I$ to be the inlier set of a global SIME solution.

The following result gives a sufficient condition for the two criteria to produce the same consensus set.

\textit{\textbf{Proposition 2.}
Suppose that $\beta=\Phi(\tau)$, and that $\Phi$ is increasing on $[0,+\infty)$ with $\Phi(0)=0$. Let $I^*$ denote a solution of MC, and let $I^+$ denote the inlier set corresponding to a global solution of SIME. Then $I^+$ is a consensus set and $|I^+|\le |I^*|$. In particular, if $\Phi(r_i({\boldsymbol{\theta}}_{I^*}))\le \Phi(\tau)$ for all $i\in I^*$, and for any other consensus set $I^\bullet\in\digamma$ there holds
\begin{equation} \label{eq07}
|I^*|-|I^\bullet|>\frac{R(I^*)-R(I^\bullet)}{\Phi(\tau)},
\end{equation}
then $I^+=I^*$.
}

\textit{\textbf{Remark 2}.
Another special condition for $I^+=I^*$ in the case of truncated LS loss is given in Lemma 23 in \cite{51}. Since $I^*$ is a largest consensus set, we have $|I^*|\ge |I^\bullet|$ for any other consensus set $I^\bullet$. Hence, with the choice $\beta=\Phi(\tau)$, SIME yields the same inlier set as MC if the fitting residual over $I^*$ is sufficiently small (or relatively small compared with that over any other consensus set $I^\bullet$), and if $\Phi(r_i({\boldsymbol{\theta}}_{I^*}))\le \Phi(\tau)$ for all $i\in I^*$. In general, however, SIME and MC do not necessarily produce the same consensus set. Moreover, a sufficient, though not necessary, condition for $I^+\neq I^*$ is that $\Phi(r_i({\boldsymbol{\theta}}_{I^*}))>\Phi(\tau)$ for some $i\in I^*$, since in this case $I^*\notin\digamma$, and hence $I^*$ cannot be the inlier set of a global SIME solution.
}

Proof of Proposition 2 is given in Appendix \ref{proof:propo2}.

Since SIME takes the fitting residual into account in inlier selection, whereas MC does not, a natural question arises:
\begin{quote}
\textit{How do SIME and MC differ in fitting performance?}
\end{quote}
We answer this question by comparing the lowest achievable fitting residual under the two criteria.

\textit{\textbf{Theorem 3 (SIME can achieve lower fitting residual than MC).}
Let $I^+$ denote the inlier set corresponding to a global solution of SIME, and let $I^*$ denote a solution of MC. Suppose that $\Phi$ is increasing on $[0,+\infty)$ with $\Phi(0)=0$. Then, with $\beta=\Phi(\tau)$, the lowest achievable residual of SIME is no larger than that of MC:
\begin{equation} \label{eq10}
\mathop{{\rm min}}\limits_{{\boldsymbol{\theta}}\in{\mathbb R}^d, h({\boldsymbol{\theta}})=0}
\sum\limits_{i\in I^+}\Phi(r_i({\boldsymbol{\theta}}))
\le
\mathop{{\rm min}}\limits_{{\boldsymbol{\theta}}\in{\mathbb R}^d, h({\boldsymbol{\theta}})=0}
\sum\limits_{i\in I^*}\Phi(r_i({\boldsymbol{\theta}})).
\end{equation}
Furthermore, if $I^* \notin \digamma$, then the inequality in (\ref{eq10}) holds strictly.}

Proof of Theorem 3 is given in Appendix \ref{proof:thrm3}.

\textit{\textbf{Remark 3.}
This theorem highlights a fundamental difference between SIME and MC in terms of the lowest achievable fitting residual. In particular, \textbf{SIME can attain a lower fitting residual than MC because it takes the fitting residual into account during inlier selection}. However, this does not imply that SIME is universally preferable to MC. In some applications, the primary goal is to identify and remove outliers, while model estimation is of secondary importance (e.g., segmentation tasks). From this perspective, SIME should be regarded as a valid alternative to MC, which maybe preferred for the applications of robust model fitting.
}

\section{Efficient Alternating Minimization Algorithms for Solving the SIME Formulation}

For the nonconvex SIME formulation (\ref{eq_sime}), a natural optimization scheme is to update the variables ${\boldsymbol{\theta}}$ and $\bf{s}$ in an alternating manner. However, since the problem is highly nonconvex involving binary variables, the direct AM algorithm can easily get trapped in local minima. To address this problem, we additionally propose an algorithm using SDR on the binary variable $\bf{s}$.

\subsection{Direct Alternating Minimization Algorithm}
The AM algorithm solves (\ref{eq_sime}) via alternatingly updating the model parameter ${\boldsymbol{\theta}}$ and the binary slack variable $\bf{s}$.
First, fixing ${\boldsymbol{\theta}}$, the $\bf{s}$-subproblem can be solved explicitly as
\begin{equation}\label{eq_am_s}
{s_i} = \left\{\! {\begin{array}{*{20}{l}}
\!{1,}&{\Phi ({r_i}({\bf{\theta }})) > \beta }\\
\!{0,}&{\Phi ({r_i}({\bf{\theta }})) \le \beta }
\end{array}} \right.\!, ~~~~ 1 \le i \le N.
\end{equation}
Then, fixing $\bf{s}$, the ${\boldsymbol{\theta}}$-subproblem is given by
\begin{equation}\label{eq_am_theta}
\mathop {{\text{min }}}\limits_{{\boldsymbol{\theta}} \in {\mathbb{R}^d}} \sum\nolimits_{i = 1}^N {\left( {1\! -\! {s_{i}}} \right)\Phi ({r_i}({\boldsymbol{\theta}}))},
~~~{\mathrm{s.t.}}~~~ h(\boldsymbol{\theta})=0,
\end{equation}
which depends on the residual models of specific applications. We consider two main cases, linear or nonlinear residual model.

For a linear residual model, which typically has a form of
\begin{equation}\label{eq18}
{r_i}({\boldsymbol{\theta}}) = | {{\mathbf{a}}_i^T{\boldsymbol{\theta}} - {b_i}} |.
\end{equation}
In robust linear regression and many computer vision applications, the residual ${r_i}({\boldsymbol{\theta}})$ can be conveniently expressed in this form. With linear residual model, if $\Phi$ is chosen as the $\ell_p$ loss, the model fitting term in SIME becomes
\begin{equation}\label{eq19}
\Phi ({r_i}({\boldsymbol{\theta}})) = {|{\mathbf{a}}_i^T{\boldsymbol{\theta}} - {b_i}|^p}.
\end{equation}
For Gaussian inlier noise, $p=2$ is the optimal choice, while for super-Gaussian inlier noise, the optimal value of $p$ should be $p<2$. Generally, a smaller $p$ should be used when the inlier noise distribution has a thicker tail.  Since the outliers are modeled as a wide uniform distribution (Proposition 1), it is reasonable to assume that the inlier noise is not too impulsive and hence we can restrict $p\geq1$, in which case $\Phi$ is convex. When $p=2$, the ${\boldsymbol{\theta}}$-subproblem with or without the constraint $\|{\boldsymbol{\theta}}\|=1$ can be solved explicitly. When $p\neq2$, it can be solved by the reweighted LS algorithm, which is detailed in Appendix \ref{Lp loss}.
The algorithm is summarized as Algorithm 1.

\begin{table}[!t]
\begin{tabular}{p{8.4cm}}
\toprule
\textbf{Algorithm 1:} {AM Algorithm}\\
\midrule
\hangafter 1
\hangindent 1.5em
\noindent
\textbf{Input:} A start point $({{\boldsymbol{\theta}}_0},{{\mathbf{s}}_0})$, and set $\beta> 0$.\\
\textbf{While} not converged ($k=0,1,2,\cdots$) \textbf{do} \\
~~~~Update ${{\mathbf{s}}}$ by (\ref{eq_am_s}) for fixed ${{\boldsymbol{\theta}}_k}$ to obtain ${{\mathbf{s}}_{k+1} }$.\\
~~~~Update ${\boldsymbol{\theta}}$ by (\ref{eq_am_theta}) for fixed ${{\mathbf{s}}_{k + 1}}$ to obtain ${{\boldsymbol{\theta}}_{k+1} }$.\\
\textbf{End while}\\
\textbf{Output:} $({{\boldsymbol{\theta}}_{k + 1}},{{\mathbf{s}}_{k + 1}})$.\\
\bottomrule
\end{tabular}
\end{table}

\subsection{Alternating Minimization with Embedded Semidefinite Relaxation}

The direct AM algorithm is efficient,
but can easily get trapped in local minima due to the high nonconvexity of (\ref{eq_sime}),
which involves binary variables. In an attempt to alleviate this problem,
we consider a relaxation of the binary variables by SDR.
SDR has been shown to be very effective in handling combinatorial problems, and it is tighter than linear relaxation \cite{24,25}.

Let $\tilde{\mathbf{ s}} \in {\{- 1,1\} ^{N + 1}}$, ${\mathbf{S}} = \tilde{\mathbf{ s}}\tilde{\mathbf{s}}^T$, ${\Phi _i}: = \Phi ({r_i}({\boldsymbol{\theta}}))$, and
\[{\mathbf{\Lambda }} = \left[\!\!{\begin{array}{*{20}{c}}
  0&{(\beta  - {{\mathbf{\Phi }}^T})/2} \\
  {(\beta  - {\mathbf{\Phi }})/2}&{{\text{diag}}({\mathbf{\Phi }})}
\end{array}} \!\!\right],\]
with ${\mathbf{\Phi }} = {[{\Phi _1},{\Phi _2}, \cdots ,{\Phi _N}]^T}$.
Then, problem (\ref{eq_sime}) is equivalent to
\begin{equation}\label{eq11}
\begin{split}
& ~~~~~~~~~~~~~~\mathop {{\text{min }}}\limits_{{\boldsymbol{\theta}} \in {\mathbb{R}^d},{\mathbf{S}} \in {\mathbb{S}^{(N + 1)}}} {\text{tr}}({\mathbf{\Lambda S}})\\
{\rm{s.t.}}~~~&{\text{diag}}({\mathbf{S}}) \!=\! {{\mathbf{1}}_{N + 1}},~{\mathbf{S}} \succeq {\mathbf{0}},~{\text{rank}}({\mathbf{S}}) \!=\! 1,~h(\boldsymbol{\theta})\!=\!0.
\end{split}
\end{equation}
Then, dropping the rank-1 nonconvex constraint leads to a SDR of (\ref{eq_sime}) as (derived in Appendix \ref{deriv_sdr})
\begin{equation}\label{eq13}
\begin{split}
&~~~~~~~~ \mathop {{\text{min }}}\limits_{{\boldsymbol{\theta}} \in {\mathbb{R}^d},{\mathbf{S}} \in {\mathbb{S}^{(N + 1)}}} {\text{tr}}({\mathbf{\Lambda S}})\\
{\rm{s.t.}} &~~~{\text{diag}}({\mathbf{S}}) = {{\mathbf{1}}_{N + 1}},~~{\mathbf{S}} \succeq {\mathbf{0}},~~h(\boldsymbol{\theta})\!=\!0.
\end{split}
\end{equation}
The above SDR is standard and widely used in binary combinatorial optimization problems \cite{25}, except that our problem additionally involves another variable ${\boldsymbol{\theta}}$ needing to be solved simultaneously. As a consequence, unlike most existing well-studied SDR problems resulting in convex relaxed formulations, (\ref{eq13}) is nonconvex.
However, for some applications with $\Phi$ and ${r_i}$ being convex, problem (\ref{eq13}) can be biconvex.

The problem (\ref{eq13}) can be solved via alternatingly updating ${\boldsymbol{\theta}}$ and ${\bf{S}}$.
Specifically, fixing ${\bf{S}}$, problem (\ref{eq13}) becomes
\begin{equation}\label{eq14}
\begin{split}
&\mathop {{\text{min }}}\limits_{{\boldsymbol{\theta}} \in {\mathbb{R}^d}} \sum\nolimits_{i = 1}^N \!{( {1\! -\! {S_{1,i + 1}}} )\Phi ({r_i}({\boldsymbol{\theta}}))}\\
&~~~~{\mathrm{s.t.}}~~~~h({\boldsymbol{\theta}})\!=\!0,
\end{split}
\end{equation}
which is a form of (\ref{eq_am_theta}).
Fixing ${\boldsymbol{\theta}}$, the ${\bf{S}}$-update subproblem becomes
\begin{equation}\label{eq15}
\begin{split}
& \mathop {{\text{min }}}\limits_{{\mathbf{S}} \in {\mathbb{S}^{(N + 1)}}} {\text{tr}}({\mathbf{\Lambda S}})\\
{\rm{s.t.}} ~~~~&{\text{diag}}({\mathbf{S}}) = {{\mathbf{1}}_{N + 1}},~~{\mathbf{S}} \succeq {\mathbf{0}}.
\end{split}
\end{equation}
It is a standard SDP and can be solved by well-established SDP solvers, such as CVX \cite{32}.
But such solvers using prime-dual interior algorithm have a complexity of $O({(N + 1)^{4.5}})$ at the worst-case, and do not scale to moderate to large problem sizes.

In order to make the algorithm scalable to high dimension problems,
the low-rank property of ${\mathbf{S}}$ can be exploited by the Burer-Monteiro (B-M) factorization \cite{25},
e.g., using ${\mathbf{S}} = {\mathbf{R}}{{\mathbf{R}}^T}$ with ${\mathbf{R}} \in {\mathbb{R}^{(N+1) \times p}}$ to recast (\ref{eq15}) into
\begin{equation}\label{eq16}
\begin{split}
& \mathop {{\text{min }}}\limits_{{\mathbf{R}} \in {\mathbb{R}^{(N+1) \times p}}} {\text{tr}}({\mathbf{\Lambda R}}{{\mathbf{R}}^T})\\
{\rm{s.t.}} &~~~~{\text{diag}}({\mathbf{R}}{{\mathbf{R}}^T}) = {{\mathbf{1}}_{N + 1}}.
\end{split}
\end{equation}
The B-M method is especially efficient to handle SDP problems, which has been applied in robotics and vision such as \cite{56}.
With this reformulation, the parameter number is reduced from ${(N + 1)^2}$ to $p(N + 1)$.
It has been proven that there exists an optimum of (\ref{eq15}) with rank less than $\lceil {\sqrt {2N} } \rceil$ \cite{25,33},
hence using $p \!\geq\! \lceil {\sqrt {2N} } \rceil$ can guarantee that any optimum of (\ref{eq16}) is also an optimum of (\ref{eq15}).
Meanwhile, although problem (\ref{eq16}) is nonconvex, it almost never has any spurious local optima \cite{34}.

\textit{\textbf{Proposition 4 \cite{34}.} For almost all ${\mathbf{\Lambda }}$, if $p(p + 1) \geq 2(N + 1)$, any local optimum ${{\mathbf{R}}^ \bullet }$ of (\ref{eq16}) is a global optimum of (\ref{eq16}), and ${{\mathbf{R}}^ \bullet }{{\mathbf{R}}^{ \bullet T}}$ is a global optimum of (\ref{eq15}).}

This result implies that, despite the nonconvexity of (\ref{eq16}), local optimization algorithms can converge to global optima. Accordingly, the first-order augmented Lagrangian algorithm \cite{25} can be used. Rather than directly using this algorithm, we exploit the sparsity structure of the problem to achieve further acceleration. This is based on the fact that only the first column, first row and diagonal of ${\mathbf{\Lambda }}$ have nonzero elements. Specifically, let ${\mathbf{r}}_i^T: = {\mathbf{R}}(i,:)$ denote the $i$-th row of ${\mathbf{R}}$ and ${\mathbf{r}}: = {[{\mathbf{r}}_1^T,{\mathbf{r}}_2^T, \cdots, {\mathbf{r}}_{N + 1}^T]^T} \in {\mathbb{R}^{p(N + 1)}}$, using the equivalence between ${\text{diag}}({\mathbf{R}}{{\mathbf{R}}^T}) = {{\mathbf{1}}_{N + 1}}$ and ${\left\| {{{\mathbf{r}}_i}} \right\|^2} = 1$ for $1 \leq i \leq N + 1$, problem (\ref{eq16}) can be reformulated as an unconstrained problem
\begin{equation}\label{eq17}
\mathop {{\text{min }}}\limits_{{\mathbf{r}} \in {\mathbb{R}^{p(N + 1)}}} J({\mathbf{r}}): = 2\sum\nolimits_{i = 2}^{N + {\text{1}}} {{\Lambda _{1,i}}\frac{{{\mathbf{r}}_1^T{{\mathbf{r}}_i}}}{{\left\| {{{\mathbf{r}}_1}} \right\|\left\| {{{\mathbf{r}}_i}} \right\|}}}
\end{equation}
where the norm-one constraints are removed by changing variable to ${[{\mathbf{R}}(i,:)]^T} = {{\mathbf{r}}_i}/\left\| {{{\mathbf{r}}_i}} \right\|$ \cite{35}, which leads to an unconstrained formulation.

Due to the equivalence between (\ref{eq16}) and (\ref{eq17}), and from Proposition 4, a local optimization algorithm can be employed to solve (\ref{eq17}), and any solution satisfies first- and second-order necessary optimality conditions is a global optimum. Hence, any efficient first-order algorithm can be employed, e.g., the limited memory BFGS (L-BFGS) algorithm \cite{36}. Such algorithms only require evaluating the first-order gradient of the objective, which is
\[{\nabla _{\mathbf{r}}}J({\mathbf{r}}) = {\left[ {{\nabla _{{\mathbf{r}}_1^T}}J({\mathbf{r}}),{\nabla _{{\mathbf{r}}_2^T}}J({\mathbf{r}}), \cdots ,{\nabla _{{\mathbf{r}}_{N + 1}^T}}J({\mathbf{r}})} \right]^T},\]
with
\begin{align}\notag
{\nabla _{{\mathbf{r}}_1}}J({\mathbf{r}}) &= 2\sum\nolimits_{i = 2}^{N + 1} {{\Lambda _{1,i}}\frac{{{{\left\| {{{\mathbf{r}}_1}} \right\|}^2}{{\mathbf{r}}_i} - {\mathbf{r}}_1^T{{\mathbf{r}}_i}{{\mathbf{r}}_1}}}{{{{\left\| {{{\mathbf{r}}_1}} \right\|}^3}\left\| {{{\mathbf{r}}_i}} \right\|}}}, \\\notag
{\nabla _{{{\mathbf{r}}_i}}}J({\mathbf{r}}) &= 2{\Lambda _{1,i}}\frac{{{{\left\| {{{\mathbf{r}}_i}} \right\|}^2}{{\mathbf{r}}_1} - {\mathbf{r}}_1^T{{\mathbf{r}}_i}{{\mathbf{r}}_i}}}{{\left\| {{{\mathbf{r}}_1}} \right\|{{\left\| {{{\mathbf{r}}_i}} \right\|}^3}}}, ~~~~{\rm{for}} ~2 \leq i \leq N + 1.
\end{align}

\begin{table}[!t]
\begin{tabular}{p{8.4cm}}
\toprule
\textbf{Algorithm 2:} {AM Algorithm with SDR (AM-R)}\\
\midrule
\hangafter 1
\hangindent 1.5em
\noindent
\textbf{Input:} A start point $({{\boldsymbol{\theta}}_0},{{\mathbf{S}}_0})$, and set $\beta> 0$.\\
\textbf{While} not converged ($k=0,1,2,\cdots$) \textbf{do} \\
~~~~Update ${{\mathbf{S}}}$ by (\ref{eq15}) for fixed ${{\boldsymbol{\theta}}_k}$ to obtain ${{\mathbf{S}}_{k+1} }$.\\
~~~~Update ${\boldsymbol{\theta}}$ by (\ref{eq14}) for fixed ${{\mathbf{S}}_{k + 1}}$ to obtain ${{\boldsymbol{\theta}}_{k+1} }$.\\
\textbf{End while}\\
\textbf{Output:} $({{\boldsymbol{\theta}}_{k + 1}},{{\mathbf{S}}_{k + 1}})$.\\
\bottomrule
\end{tabular}
\end{table}

\section{Application To 3D Registration}

This section presents the application of SIME to rotation search and 6-DoF rigid registration. Rotation search, also known as the Wahba problem, aims to estimate the rotation between two coordinate frames, which has wide applications in computer vision, robotics, and aerospace engineering \cite{42,43,44,45,46}. 6-DoF rigid registration can be viewed as an extension of 3-DoF rotation search, which estimates both rotation and translation.

\subsection{Rotation Search}

Consider a set of 3D point pairs $\{ ({{\mathbf{a}}_i}{\text{,}}{{\mathbf{b}}_i}):i = 1,2, \cdots, N\}$ with ${{\mathbf{a}}_i},{{\mathbf{b}}_i} \in {\mathbb{R}^3}$, which are generated as
\begin{equation}\label{eq24}
{{\mathbf{b}}_i} = {\mathbf{R}}{{\mathbf{a}}_i} + {{\mathbf{n}}_i} + {{\mathbf{o}}_i},
\end{equation}
where ${\mathbf{R}} \in {\text{SO}}(3)$ is the unknown rotation, ${{\mathbf{n}}_i}$ models small inlier measurement noise, ${{\mathbf{o}}_i}$ is zero if the data pair ${\text{(}}{{\mathbf{a}}_i}{\text{,}}{{\mathbf{b}}_i})$ is inlier, or ${{\mathbf{o}}_i}$ is an arbitrary perturbation if ${\text{(}}{{\mathbf{a}}_i}{\text{,}}{{\mathbf{b}}_i})$ is outlier.

For convenience, we adopt quaternion representation for 3D rotation \cite{47}. Denote a unit quaternion by ${\mathbf{q}} = [{{\mathbf{v}}^T}~s]^T$, where ${\mathbf{v}} \in {\mathbb{R}^3}$ is the vector part and $s$ is the scalar part. If ${\mathbf{R}}$ is the unique rotation corresponding to a unit quaternion ${\mathbf{q}}$, then the rotation of a vector ${\mathbf{a}} \in {\mathbb{R}^3}$ by ${\mathbf{R}}$ can  be expressed in terms of quaternion product as
\[\left[\!\! {\begin{array}{*{20}{c}}
  {{\mathbf{Ra}}} \\
  {\text{0}}
\end{array}}\!\! \right] = {\mathbf{q}} \circ \hat{\mathbf{ a}} \circ {{\mathbf{q}}^{ - 1}},\]
where ${{\mathbf{q}}^{ - 1}} = {[ - {{\mathbf{v}}^T}~s]^T}$ is the quaternion inverse, and ${\mathbf{\hat a}} = {[{{\mathbf{a}}^T}~0]^T}$. The quaternion product is defined as ${\mathbf{q}} \circ {\mathbf{x}} = \Omega ({\mathbf{q}}){\mathbf{x}}$ for any ${\mathbf{x}} \in {\mathbb{R}^4}$, and ${{\mathbf{q}}_1} \circ {{\mathbf{q}}_2} = \Omega ({{\mathbf{q}}_1}){{\mathbf{q}}_2} = \bar \Omega ({{\mathbf{q}}_2}){{\mathbf{q}}_1}$ for two unit quaternion ${{\mathbf{q}}_1}$ and ${{\mathbf{q}}_2}$, where
\[\Omega ({\mathbf{q}}) \!\!=\!\! \left[\!\!\! {\begin{array}{*{20}{c}}
  {{q_4}}\!&\!{ - {q_3}}\!&\!{{q_2}}\!&\!{{q_1}} \\
  {{q_3}}\!&\!{{q_4}}\!&\!{ - {q_1}}\!&\!{{q_2}} \\
  { - {q_2}}\!&\!{{q_1}}\!&\!{{q_4}}\!&\!{{q_3}} \\
  { - {q_1}}\!&\!{ - {q_2}}\!&\!{ - {q_3}}\!&\!{{q_4}}
\end{array}} \!\!\!\right]\!\!,
~~~\bar \Omega ({\mathbf{q}}) \!\!=\!\! \left[\!\!\! {\begin{array}{*{20}{c}}
  \!{{q_4}}\!&\!{{q_3}}\!&\!{ - {q_2}}\!&\!{{q_1}} \\
  \!{ - {q_3}}\!&\!{{q_4}}\!&\!{{q_1}}\!&\!{{q_2}} \\
  \!{{q_2}}\!&\!{ - {q_1}}\!&\!{{q_4}}\!&\!{{q_3}} \\
 \! { - {q_1}}\!&\!{ - {q_2}}\!&\!{ - {q_3}}\!&\!{{q_4}}
\end{array}} \!\!\!\right]\!\!.\]

Based on quaternion representation, the linear residual with a LS loss can be expressed as
\begin{equation}\label{eq25}
\Phi ({r_i}({\boldsymbol{\theta}})) = {\| {{{\hat{\mathbf{ b}}}_i} - {\boldsymbol{\theta}} \circ {{\hat{\mathbf{ a}}}_i} \circ {{\boldsymbol{\theta}}^{ - 1}}} \|^2}, ~~{\rm{with}}~~\left\| {\boldsymbol{\theta}} \right\| = 1.
\end{equation}
For the AM-R algorithm, similar to (\ref{eq13}), the SDR of SIME in this case leads to
\begin{equation}\label{eq26}
\begin{split}
& \mathop {{\text{min }}}\limits_{{\boldsymbol{\theta}} \in {\mathbb{R}^4},{\mathbf{S}} \in {\mathbb{S}^{(N + 1)}}} {\text{tr}}({\mathbf{\Lambda S}})\\
{\rm{s.t.}} &~~{\text{diag}}({\mathbf{S}}) = {{\mathbf{1}}_{N + 1}},~{\mathbf{S}} \succeq {\mathbf{0}},~\left\| {\boldsymbol{\theta}} \right\| \!= \!1.
\end{split}
\end{equation}
Accordingly, the ${\boldsymbol{\theta}}$-subproblem becomes
\begin{equation}\label{eq27}
\begin{split}
&\mathop {{\text{min }}}\limits_{{\boldsymbol{\theta}} \in {\mathbb{R}^4}} \sum\limits_{i = 1}^N {\left( {1 - {S_{1,i + 1}}} \right){{\left\| {{{\hat{\mathbf{ b}}}_i} - {\boldsymbol{\theta}} \circ {{\hat{\mathbf{ a}}}_i} \circ {{\boldsymbol{\theta}}^{ - 1}}} \right\|}^2}}\\
&~~~~{\rm{s.t.}}~~\left\| {\boldsymbol{\theta}} \right\| = 1.
\end{split}
\end{equation}
For a unit quaternion ${\boldsymbol{\theta}}$, it follows that $c = {{\boldsymbol{\theta}}^T}(c{{\mathbf{I}}_{4}}){\boldsymbol{\theta}}$ for any $c \in \mathbb{R}$, $\hat{\mathbf{ b}}_i^T( {{\boldsymbol{\theta}} \circ {{\hat{\mathbf{ a}}}_i} \circ {{\boldsymbol{\theta}}^{ - 1}}} ) = {{\boldsymbol{\theta}}^T}{\Omega ^T}({\hat{\mathbf{ b}}_i})\bar \Omega ({\hat{\mathbf{ a}}_i}){\boldsymbol{\theta}}$ and $ - {\Omega ^T}({\hat{\mathbf{ b}}_i}) = \Omega ({\hat{\mathbf{ b}}_i})$, hence problem (\ref{eq27}) can be rewritten as
\begin{equation}\label{eq28}
\mathop {{\text{min }}}\limits_{{\boldsymbol{\theta}} \in {\mathbb{R}^4}} {{\boldsymbol{\theta}}^T}{\mathbf{G}}{\boldsymbol{\theta }},~~~~
{\rm{s.t.}}~~\left\| {\boldsymbol{\theta}} \right\| = 1,
\end{equation}
with
\[{\mathbf{G}} = \sum\limits_{i = 1}^N {\left( {1 - {\mathbf{S}}_{1,i + 1}} \right)\left[ {\left( {{{\left\| {{{\mathbf{b}}_i}} \right\|}^2} + {{\left\| {{{\mathbf{a}}_i}} \right\|}^2}} \right){{\mathbf{I}}_{\text{4}}} + 2\Omega ({{\hat{\mathbf{ b}}}_i})\bar \Omega ({{\hat{\mathbf{ a}}}_i})} \right]}. \]
Obviously, the solution of (\ref{eq28}) is given by the eigenvector corresponding to the smallest eigenvalue of ${\mathbf{G}}$.

Next, we compare the proposed method with a close existing method \cite{31},
namely QUASAR (QUAternion-based Semidefinite relAxation for Robust alignment).
QUASAR uses a truncated LS loss and has a formulation as
\begin{equation}\label{eq29}
\mathop {{\text{min }}}\limits_{\begin{subarray}{l}
  {\boldsymbol{\theta}} \in {\mathbb{R}^4},\left\| {\boldsymbol{\theta}} \right\| = 1 \\
  {s_i} \in \{  - 1,1\}
\end{subarray}}  \sum\limits_{i = 1}^N {\frac{{1 - {s_i}}}{2}\frac{{{{\left\| {{{\hat{\mathbf{ b}}}_i} - {\boldsymbol{\theta}} \circ {{\hat{\mathbf{ a}}}_i} \circ {{\boldsymbol{\theta}}^{ - 1}}} \right\|}^2}}}{{{\sigma ^2}}} + } \frac{{1 + {s_i}}}{2}{\bar c^2},
\end{equation}
which can be viewed as a special instance of the SIME formulation.
To solve this mixed-integer program, QUASAR adopts a binary cloning based reformulation as
\begin{equation}\label{eq30}
\begin{split}
\mathop {{\text{min }}}\limits_{\begin{subarray}{l}
  {\boldsymbol{\theta}} \in {\mathbb{R}^4},\left\| {\boldsymbol{\theta}} \right\| = 1 \\
  {{\boldsymbol{\theta}}_i} =  \pm {\boldsymbol{\theta}}
\end{subarray}}  \sum\limits_{i = 1}^N &{\frac{{{{\left\| {{{\hat{\mathbf{ b}}}_i} \!-\! {\boldsymbol{\theta}}\! \circ\! {{\hat{\mathbf{ a}}}_i} \!\circ\! {{\boldsymbol{\theta}}^{ - 1}}\! -\! {{\boldsymbol{\theta}}^T}{{\boldsymbol{\theta}}_i}{{\hat{\mathbf{ b}}}_i} \!+\! {\boldsymbol{\theta}}\! \circ\! {{\hat{\mathbf{ a}}}_i} \!\circ\! {\boldsymbol{\theta}}_i^{ - 1}} \right\|}^2}}}{{4{\sigma ^2}}}}  \! \\
&~~~~~~~~~~~~~~~~~~~~~~+ \frac{{1 + {{\boldsymbol{\theta}}^T}{{\boldsymbol{\theta}}_i}}}{2}{{\bar c}^2}.  \\
\end{split}
\end{equation}
This reformulation is based on the fact that, if ${{\boldsymbol{\theta}}_i} = {s_i}{\boldsymbol{\theta}}$ with ${s_i} \in \{  - 1,1\} $, then ${{\boldsymbol{\theta}}^T}{{\boldsymbol{\theta}}_i} = {s_i}$ and ${\boldsymbol{\theta}} \circ {\hat{\mathbf{ a}}_i} \circ {\boldsymbol{\theta}}_i^{ - 1} = {s_i}({\boldsymbol{\theta}} \circ {\hat{\mathbf{ a}}_i} \circ {{\boldsymbol{\theta}}^{ - 1}})$. Let $\tilde{\boldsymbol{ \theta }} = [{\boldsymbol{\theta}}^T,{\boldsymbol{\theta}}_1^T, \cdots ,{\boldsymbol{\theta}}_N^T]^T$, problem (\ref{eq30}) can be expressed as \cite{31}
\begin{equation}\label{eq31}
\begin{split}
&\mathop {{\text{min}}}\limits_{\tilde{\boldsymbol{\theta}} \in {\mathbb{R}^{4(N + 1)}}} {\text{ }}{\tilde{\boldsymbol{\theta}}^T}{\mathbf{Q}}\tilde {\boldsymbol{\theta}}\\
{\rm{s.t.}}~~&\left\| {\boldsymbol{\theta}} \right\| = 1, ~{{\boldsymbol{\theta}}_i}{\boldsymbol{\theta}}_i^T = {\boldsymbol{\theta}}{{\boldsymbol{\theta}}^T},~\forall i = 1, \cdots ,N,
\end{split}
\end{equation}
where ${\mathbf{Q}} \in {\mathbb{R}^{4(N + 1) \times 4(N + 1)}}$ is given by
\begin{equation}\label{eq32}\notag
{\mathbf{Q}} = \left[\!\! {\begin{array}{*{20}{c}}
  {\mathbf{0}}&{{{\mathbf{Q}}_{01}}}& \cdots &{{{\mathbf{Q}}_{0N}}} \\
  {{{\mathbf{Q}}_{01}}}&{{{\mathbf{Q}}_{11}}}& \cdots &{\mathbf{0}} \\
   \vdots & \vdots & \ddots & \vdots  \\
  {{{\mathbf{Q}}_{0N}}}&{\mathbf{0}}& \cdots &{{{\mathbf{Q}}_{NN}}}
\end{array}} \!\!\right],
\end{equation}
with
\[{{\mathbf{Q}}_{ii}} = \frac{{({{\left\| {{{\mathbf{b}}_i}} \right\|}^2} + {{\left\| {{{\mathbf{a}}_i}} \right\|}^2}){{\mathbf{I}}_{\text{4}}} + 2\Omega ({{\hat{\mathbf{ b}}}_i})\bar \Omega ({{\hat{\mathbf{ a}}}_i})}}{{2{\sigma ^2}}} + \frac{{{{\bar c}^2}}}{2}{{\mathbf{I}}_{\text{4}}},\]
\[{{\mathbf{Q}}_{0i}} =  - \frac{{({{\left\| {{{\mathbf{b}}_i}} \right\|}^2} + {{\left\| {{{\mathbf{a}}_i}} \right\|}^2}){{\mathbf{I}}_{\text{4}}} + 2\Omega ({{\hat{\mathbf{ b}}}_i})\bar \Omega ({{\hat{\mathbf{ a}}}_i})}}{{4{\sigma ^2}}} + \frac{{{{\bar c}^2}}}{4}{{\mathbf{I}}_{\text{4}}}.\]
Then, let ${\mathbf{Z}} = \tilde{\boldsymbol{\theta}}{\tilde{\boldsymbol{\theta}}^T} \in {\mathbb{S}^{4(N + 1)}}$ and denote its $4 \times 4$ sub-blocks by ${[{\mathbf{Z}}]_{ij}} = {{\boldsymbol{\theta}}_i}{\boldsymbol{\theta}}_j^T$ for $\forall 0 \leq i,j \leq N$ with ${{\boldsymbol{\theta}}_0} = {\boldsymbol{\theta}}$, QUASAR adopts a SDR of (\ref{eq31}) as
\begin{equation}\label{eq33}
\begin{split}
&\mathop {{\text{min}}}\limits_{{\mathbf{Z}} \in {\mathbb{S}^{4(N + 1)}}} {\text{tr}}({\mathbf{QZ}})\\
{\rm{s.t.}}~~&{\text{tr}}({[{\mathbf{Z}}]_{00}}) = 1,~{\mathbf{Z}} \succeq {\mathbf{0}},\\
&{[{\mathbf{Z}}]_{ii}} = {[{\mathbf{Z}}]_{00}},~\forall i = 1, \cdots, N,\\
&{[{\mathbf{Z}}]_{ij}} = [{\mathbf{Z}}]_{ij}^T,~\forall 0 \leq i < j \leq N.~~~~
\end{split}
\end{equation}

The next result compares QUASAR with the proposed AM-R algorithm.

\textit{\textbf{Proposition 5.} If the SIME formulation uses the loss (\ref{eq25}) and with $\beta  = {\sigma ^2}{\bar c^2}$, then it is equivalent to the QUASAR formulations (\ref{eq29})--(\ref{eq31}). Furthermore, the relaxation (\ref{eq26}) of SIME is tighter than the relaxation (\ref{eq33}) of QUASAR.}

\begin{proof} 
When the loss (\ref{eq25}) and $\beta = {\sigma ^2}{\bar c^2}$ are used in SIME, it is easy to see the equivalence between SIME and (\ref{eq29}) by changing variable from ${s_i} \in \{ 0,1\} $ to ${s_i} \in \{- 1,1\} $. To justify (\ref{eq26}) is tighter than (\ref{eq33}), we first show that
\begin{equation}\label{eq34}
{\text{tr}}({\mathbf{\Lambda S}}) = 2{\sigma ^2}{\text{tr}}\left( {{\mathbf{Q}}\left( {{\mathbf{S}} \otimes ({\boldsymbol{\theta}}{{\boldsymbol{\theta}}^T})} \right)} \right) - {\sigma ^2}{\bar c^2}.
\end{equation}
From the properties of the trace and Kronecker product operations, some algebra leads to ${\text{tr}}\left( {{\mathbf{Q}}\left( {{\mathbf{S}} \otimes ({\boldsymbol{\theta}}{{\boldsymbol{\theta}}^T})} \right)} \right) = {{\boldsymbol{\theta}}^T}\bar{\mathbf{ Q}}{\boldsymbol{\theta }}$ with
\[\bar{\mathbf{ Q}} = \sum\nolimits_{i = 1}^N {{{\mathbf{Q}}_{ii}} + 2{S_{1,i + 1}}{{\mathbf{Q}}_{0i}}}. \]
Similarly, with the loss (\ref{eq25}) and $\beta = {\sigma ^2}{\bar c^2}$, it follows that ${\text{tr}}({\mathbf{\Lambda S}}) = 2{\sigma ^2}{{\boldsymbol{\theta}}^T}\bar{\mathbf{ Q}}{\boldsymbol{\theta }} - {\sigma ^2}{\bar c^2}$, which leads to (\ref{eq34}). Hence, the relaxation (\ref{eq26}) is equivalent to
\begin{equation}\notag
\begin{split}
&\mathop {{\text{min}}}\limits_{{\boldsymbol{\theta}} \in {\mathbb{R}^{\text{4}}},{\mathbf{S}} \in {\mathbb{S}^{(N + 1)}}} {\text{ tr}}\left( {{\mathbf{Q}}\left( {{\mathbf{S}} \otimes ({\boldsymbol{\theta}}{{\boldsymbol{\theta}}^T})} \right)} \right)\\
&{\rm{s.t.}}~~\left\| {\boldsymbol{\theta}} \right\| = 1,~{\text{diag}}({\mathbf{S}}) = {{\mathbf{1}}_{N + 1}},~{\mathbf{S}} \succeq {\mathbf{0}},
\end{split}
\end{equation}
which is further equivalent to
\begin{equation}\label{eq35}
\begin{split}
&\mathop {{\text{min}}}\limits_{{\mathbf{Z}} \in {\mathbb{S}^{4(N + 1)}}} {\text{ tr}}({\mathbf{QZ}})\\
{\rm{s.t.}}&~~{[{\mathbf{Z}}]_{00}} = {\boldsymbol{\theta}}{{\boldsymbol{\theta}}^T},~\left\| {\boldsymbol{\theta}} \right\| = 1,\\
&~~{[{\mathbf{Z}}]_{ii}} = {[{\mathbf{Z}}]_{00}},~\forall i = 1, \cdots ,N,\\
&~~{[{\mathbf{Z}}]_{ij}} = {s_{ij}}{[{\mathbf{Z}}]_{00}},~\forall 0 \leq i < j \leq N,\\
&~~{\mathbf{S}} = \left[\!\! {\begin{array}{*{20}{c}}
  1&{{s_{01}}}& \cdots &{{s_{0N}}} \\
  {{s_{01}}}&1& \cdots &{{s_{1N}}} \\
   \vdots & \vdots & \ddots & \vdots  \\
  {{s_{0N}}}&{{s_{1N}}}& \cdots &1
\end{array}} \!\!\right] \succeq {\mathbf{0}}.
\end{split}
\end{equation}
Then, it is easy to see that the constraints in (\ref{eq35}) in fact constitute a subset of the constraints in (\ref{eq33}).
Hence, the optimum objective of (\ref{eq35}) (equivalently that of (\ref{eq26})) provides a tighter lower bound to the original nonconvex problem than (\ref{eq33}).
\end{proof}

\textit{\textbf{Remark 4 (Computational complexity).} It has been shown in \cite{31} that the relaxation (\ref{eq33}) with redundant constraints is sufficiently tight. Particularly, in the noiseless and outlier-free case, it is always tight as its optimal solution has rank-1 and attains a global minimum of the original nonconvex problem. However, QUASAR solving (\ref{eq33}) is computationally expensive and scales poorly in problem size. For example, with a general SDP solver it typically needs more than 1000 seconds for $N=100$ \cite{31}. Although (\ref{eq33}) is also a SDP like the ${\bf{S}}$-subproblem (\ref{eq15}), the accelerating procedure in Section 3.2 does not apply to it. That is because the SDP (\ref{eq33}) involves a large number of equality constraints, about $3{N^2} + 13N$. Meanwhile, the constraints do not admit an unconstrained formulation. Hence, when using the augmented Lagrangian method, a large number of dual variables (about $3{N^2} + 13N$) have to be handled, which fundamentally increases the computational complexity. In comparison, (\ref{eq15}) only has $N+1$ equality constraints and, more importantly, the norm-one constraints have a special ``hidden convexity'' structure admitting an unconstrained formulation (\ref{eq17}). This leads to a significant advantage of our algorithm over QUASAR in computational efficiency, e.g., three orders of magnitude faster as will be shown in experiments.}

Moreover, for the AM algorithm, the ${\boldsymbol{\theta}}$-subproblem can be expressed as (\ref{eq27}) with $({1\! -\! {S_{1,i + 1}}})$ replaced by $({1\! \!- {s_{i}}})$, where $s_i$ is computed by (7).

\subsection{6-DoF Euclidean Registration}

In the context of 6-DoF Euclidean registration defined by a rigid transformation $[{\mathbf{R}},{\mathbf{t}}] \in {\mathbb{R}^{3 \times 4}}$, where ${\mathbf{R}} \in {\text{SO}}(3)$ and ${\mathbf{t}}$ are the unknown rotation and translation, respectively, the generation model (\ref{eq24}) is extended to
\begin{equation}\label{eq36}
{{\mathbf{b}}_i} = {\mathbf{R}}{{\mathbf{a}}_i} + {\mathbf{t}} + {{\mathbf{n}}_i} + {{\mathbf{o}}_i}.
\end{equation}
In this case, using quaternion representation for rotation and with the LS loss, the fitting objective becomes
\begin{equation}\label{eq37}
\Phi ({r_i}({\boldsymbol{\theta}})) = {\left\| {{{\hat{\mathbf{ b}}}_i} - {\boldsymbol{\theta}} \circ {{\hat{\mathbf{ a}}}_i} \circ {{\boldsymbol{\theta}}^{ - 1}} - \hat{\mathbf{ t}}} \right\|^2},~~{\rm{with}}~~\left\| {\boldsymbol{\theta}} \right\| = 1,
\end{equation}
where $\hat{\mathbf{ t}} = {[{{\mathbf{t}}^T}~0]^T}$.
Accordingly, the ${\boldsymbol{\theta}}$-subproblem becomes
\begin{equation}\label{eq38}
\begin{split}
&\mathop {{\text{min }}}\limits_{{\boldsymbol{\theta}} \in {\mathbb{R}^4},{\mathbf{t}} \in {\mathbb{R}^3}} \sum\limits_{i = 1}^N {\omega_i{{\left\| {{{\hat{\mathbf{ b}}}_i} \!-\! {\boldsymbol{\theta}} \!\circ\! {{\hat{\mathbf{ a}}}_i} \!\circ\! {{\boldsymbol{\theta}}^{ - 1}} \!-\! \hat{\mathbf{ t}}} \right\|}^2}}\\
&~~~~{\rm{s.t.}}~~\left\| {\boldsymbol{\theta}} \right\| = 1.
\end{split}
\end{equation}
For the AM algorithm, $\omega_i = 1 - s_i$ with $s_i$ be computed by (\ref{eq_am_theta}),
whilst for the AM-R algorithm, $\omega_i = {1 - {S_{1,i + 1}}}$ with $S_{i,j}$ be computed by (\ref{eq15}).
Then, the closed-form solution to problem (\ref{eq38}) is given in \cite{55}.

\section{Experimental Results}

We evaluate the proposed algorithms in 3D registration experiments,
including simulated 3-DoF rotation registration and 6-DoF Euclidean registration, and a real-world 3D registration experiment.
For the AM-R algorithm (Algorithm 2), we set $p = \lceil {\sqrt {2N}/3} \rceil$ for the low-rank factorization (\ref{eq16}) and solve (\ref{eq17}) by L-BFGS using \textit{minFunc} \cite{54}.

\subsection{Rotation Registration}

\begin{figure*}[!t]
\centering
\subfigure[$N = 100$, $\sigma = 0.01$ (low noise)]{{\includegraphics[width = 8cm]{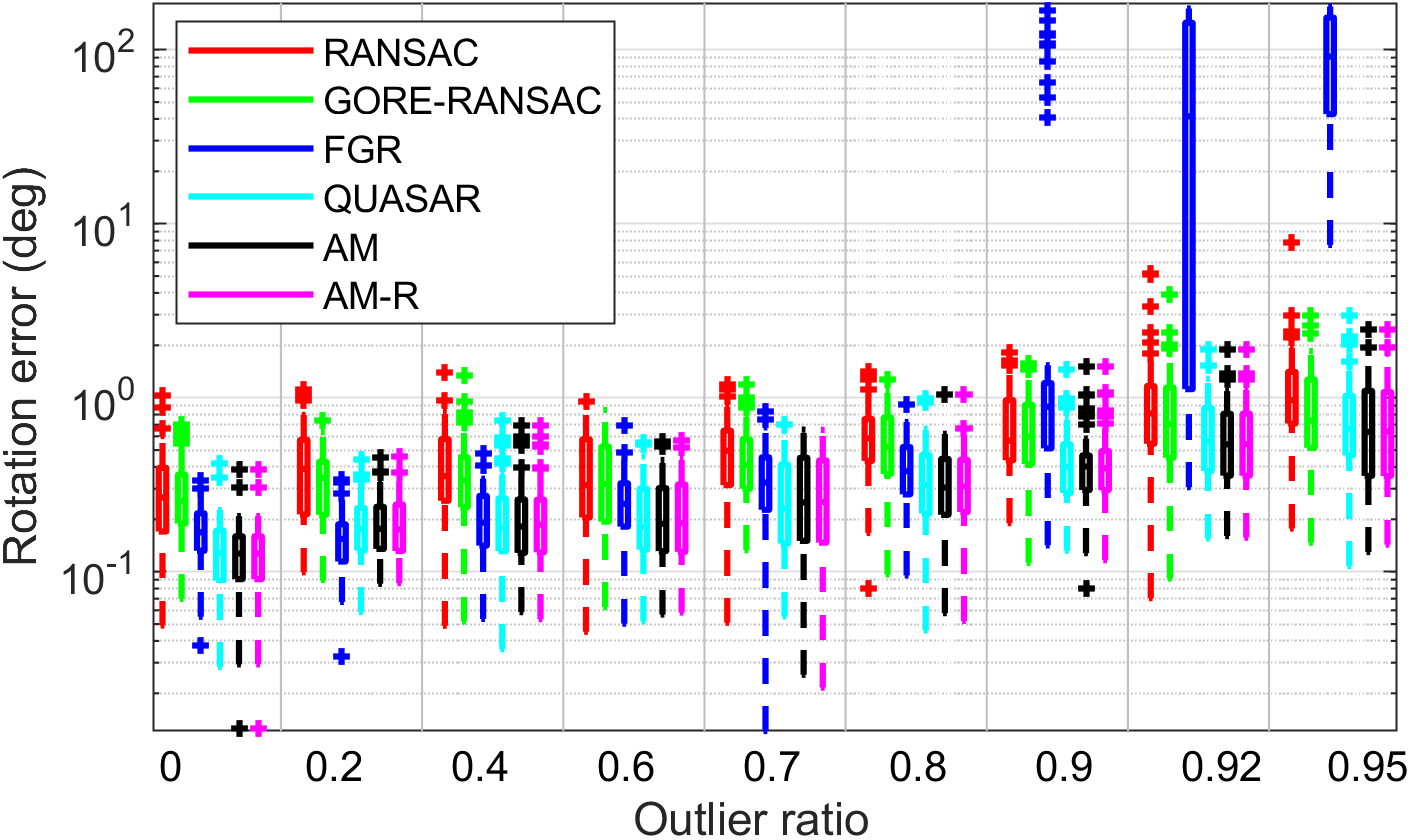}}}~~~~~~~~~~~~
\subfigure[$N = 100$, $\sigma = 0.1$ (high noise)]{{\includegraphics[width = 8cm]{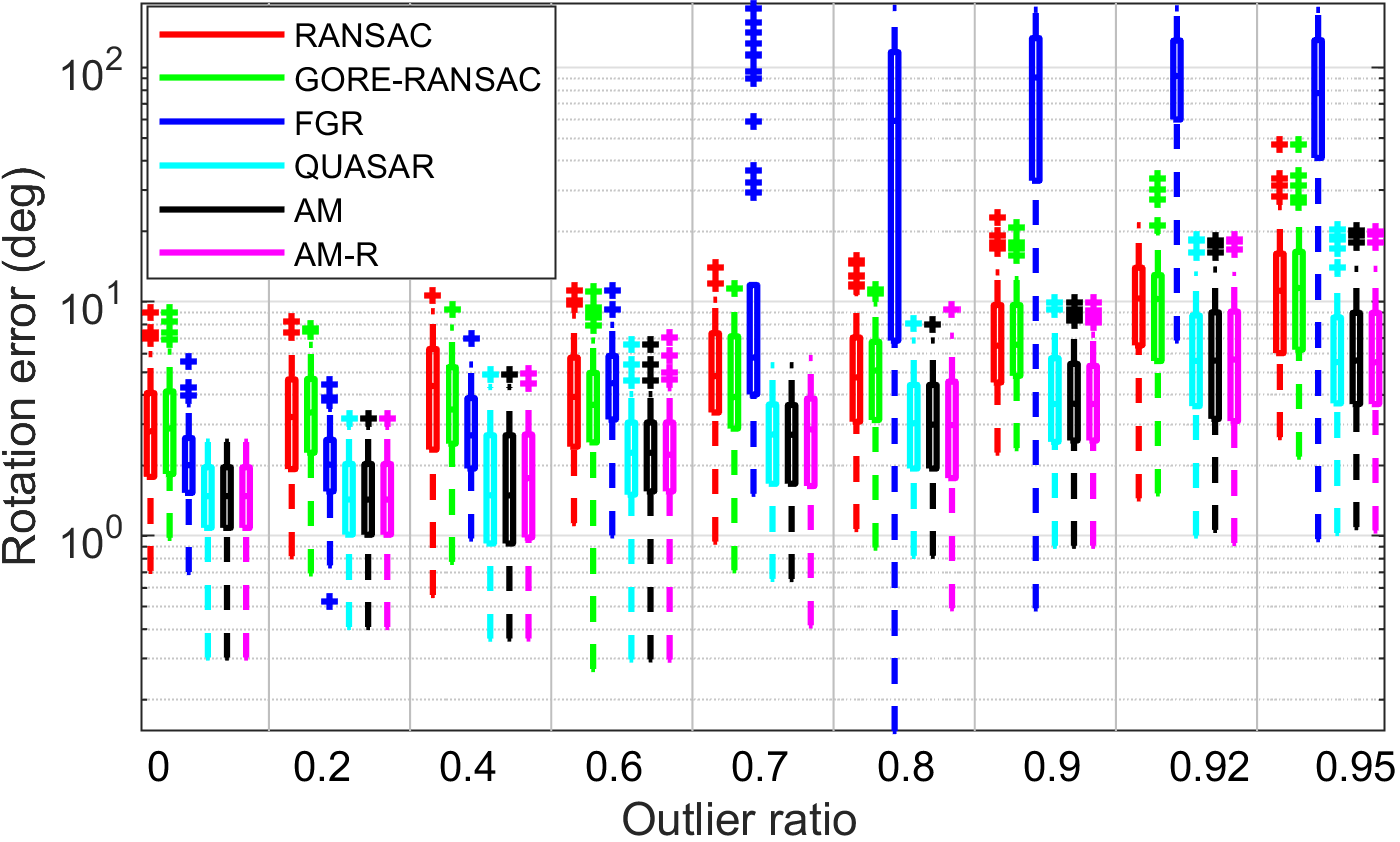}}}\\
\subfigure[$N = 500$, $\sigma = 0.01$ (low noise)]{{\includegraphics[width = 8cm]{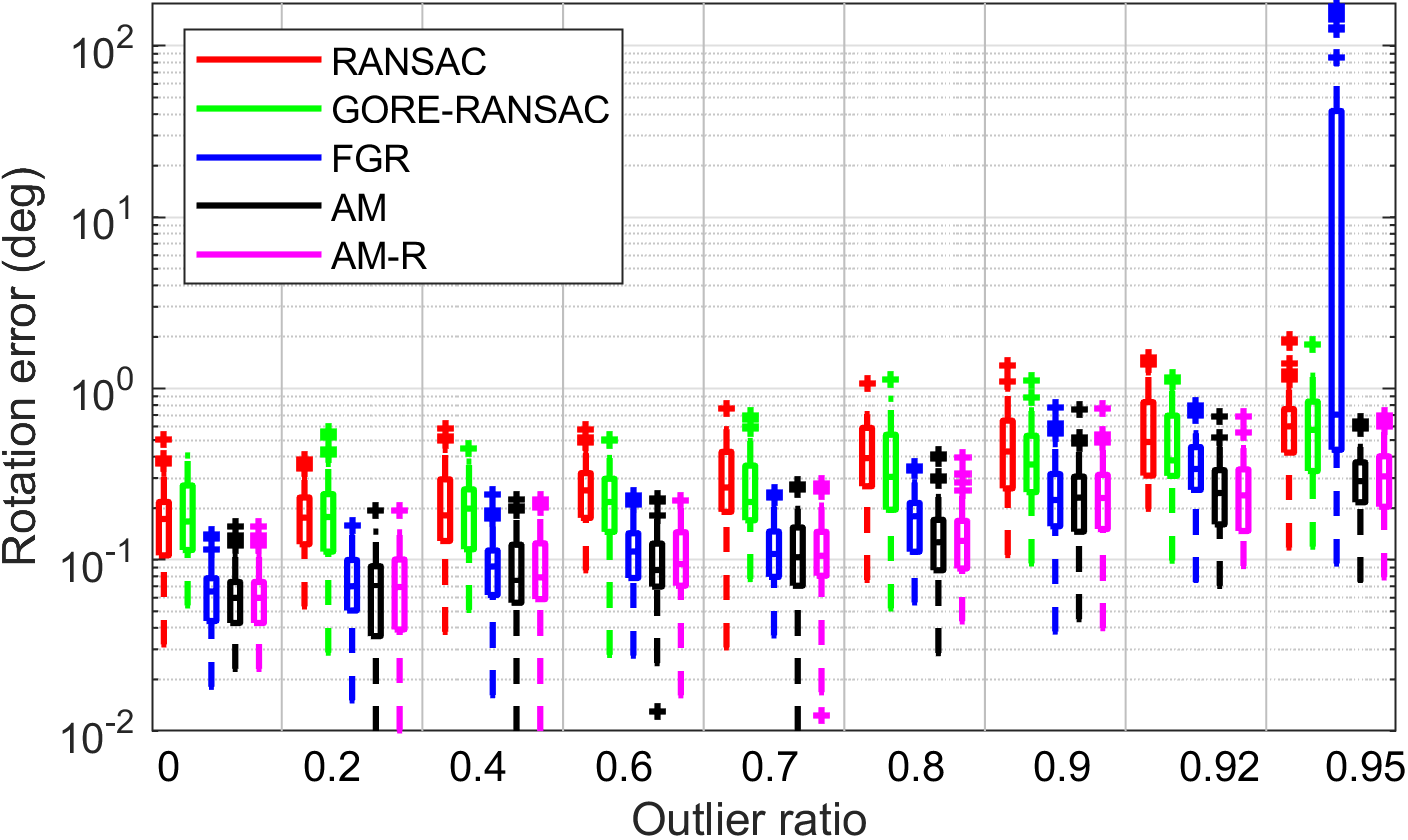}}}~~~~~~~~~~~~
\subfigure[$N = 500$, $\sigma = 0.1$ (high noise)]{{\includegraphics[width = 8cm]{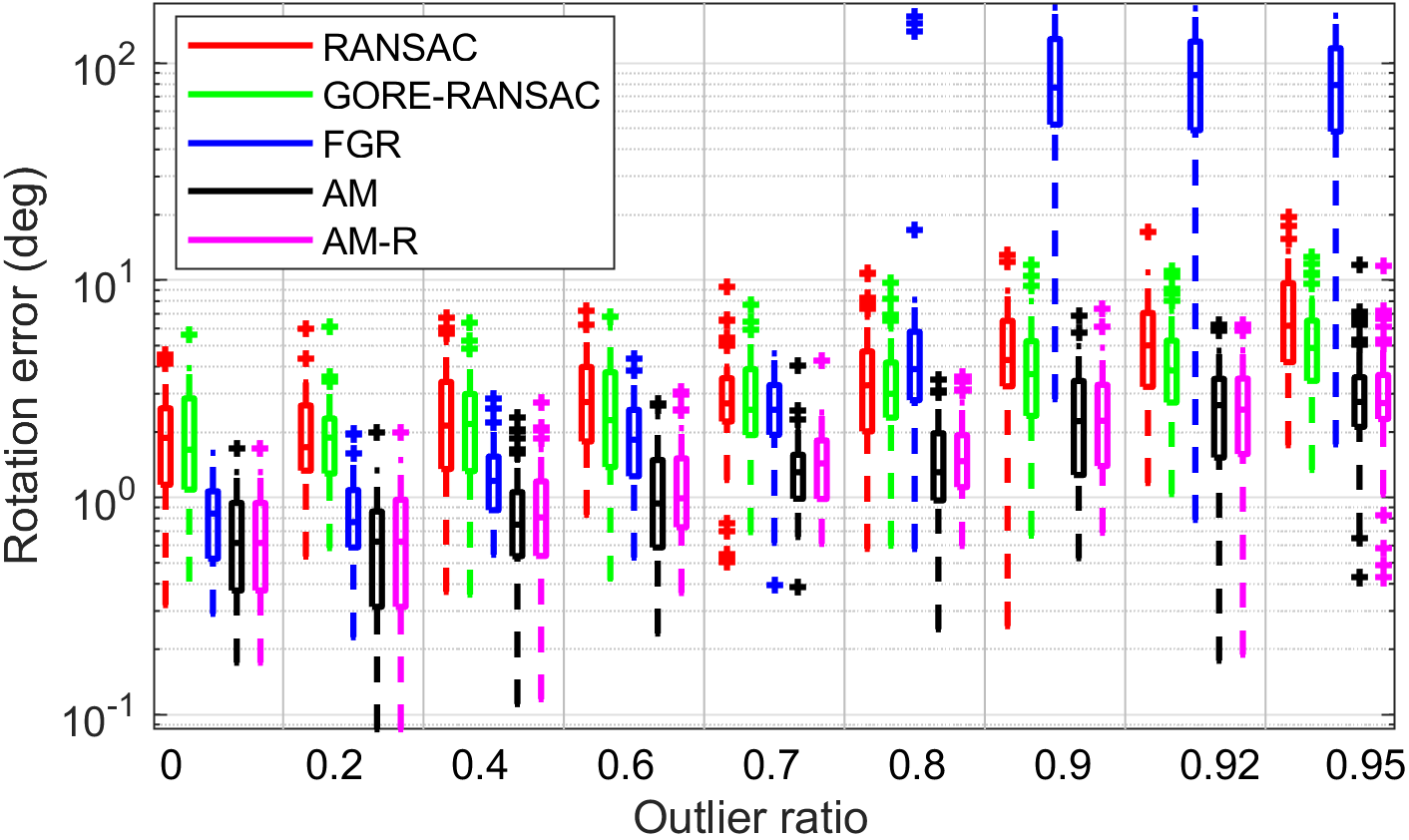}}}
\caption{Rotation error comparison in the rotation registration experiment with low and high noise conditions. (a) $N = 100$, $\sigma = 0.01$. (b) $N = 100$, $\sigma = 0.1$. (c) $N = 500$, $\sigma = 0.01$. (d) $N = 500$, $\sigma = 0.1$. QUASAR is not compared in the case of $N=500$ as it runs out of memory when $N>150$.}
\label{figure8}
\end{figure*}

\begin{figure*}[!t]
\centering
\subfigure[$N = 100$]{{\includegraphics[width = 8.3cm]{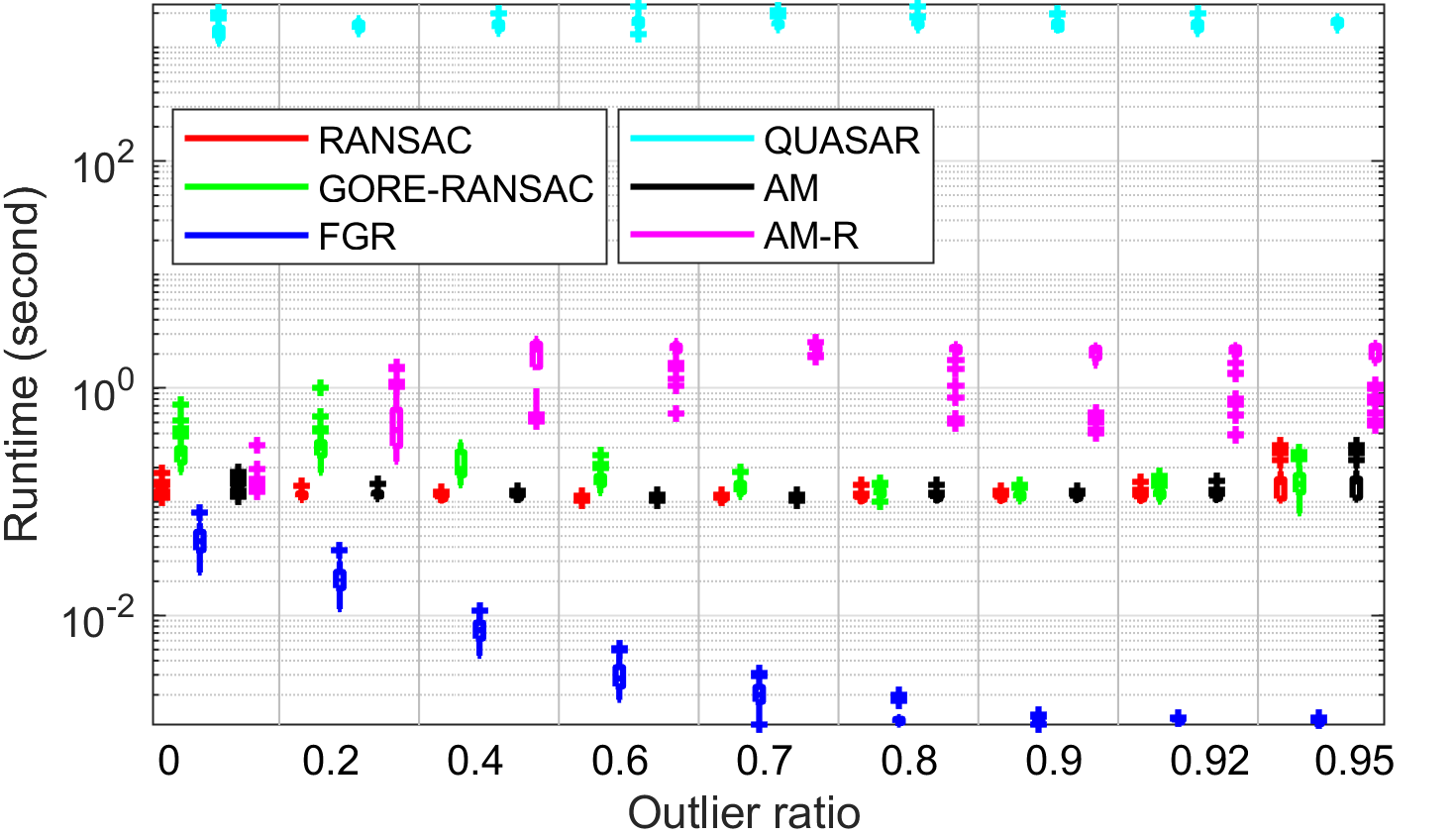}}}~~~~~~~~
\subfigure[$N = 500$]{{\includegraphics[width = 8.3cm]{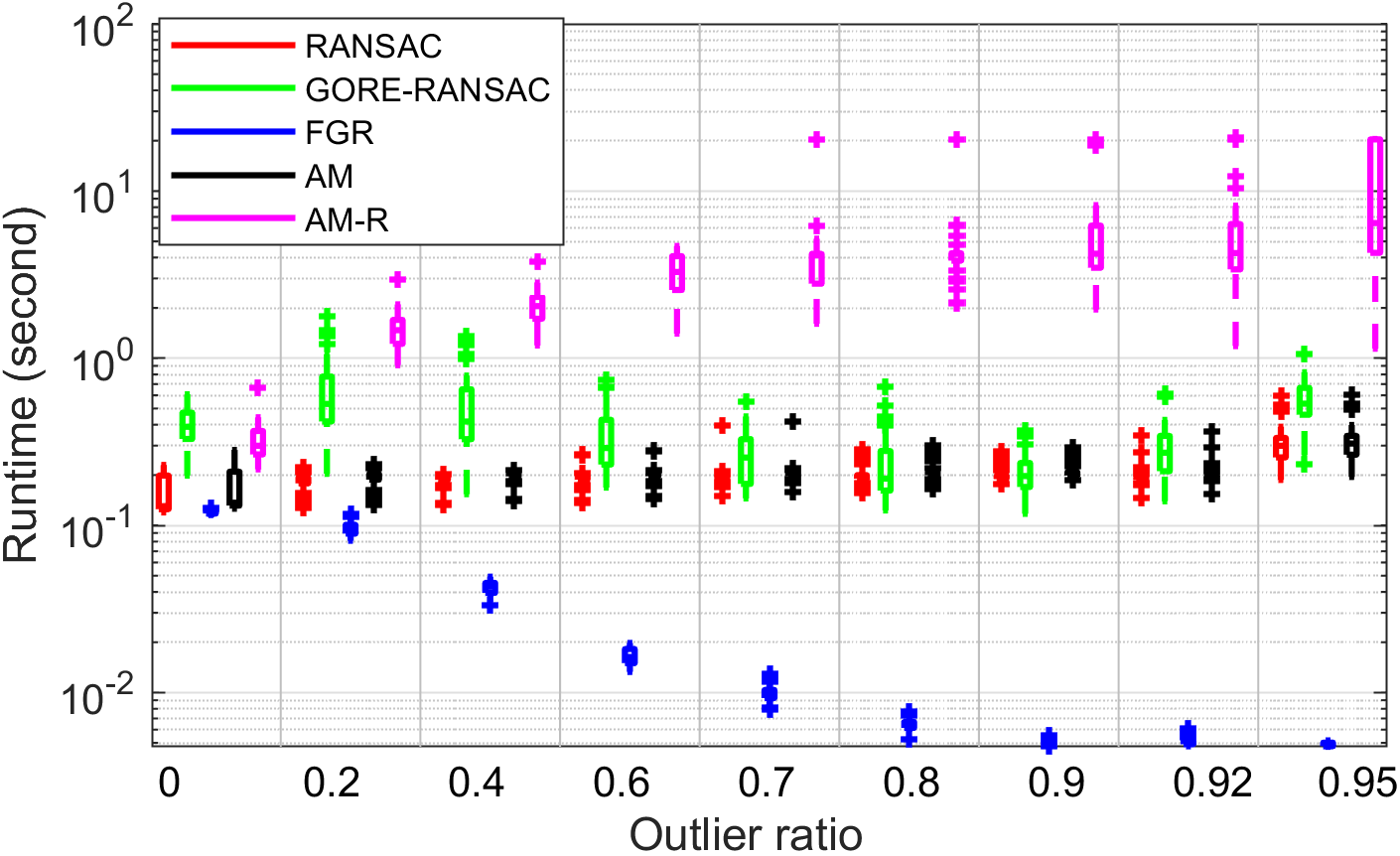}}}
\caption{Runtime comparison in the rotation registration experiment for two cases, (a) $N = 100$, and (b) $N =500$ (with $\sigma = 0.1$).}
\label{figure9}
\end{figure*}

In the first experiment, we evaluate the new algorithms on rotation search, in comparison with \textit{i)} RANSAC; \textit{ii)} GORE-RANSAC, which uses GORE \cite{50} to firstly remove most outliers and then uses RANSAC to estimate the model from the pruned measurements; \textit{iii)} Fast global registration (FGR) \cite{30}; \textit{iv)} QUASAR \cite{31}, which is implemented in Matlab using CVX \cite{32} with MOSEK \cite{52} as the SDP solver.
Note that GORE can provide a rough estimation of the model with guaranteed outlier removal, but GORE-RANSAC has significantly better accuracy in most cases.
For QUASAR, AM and AM-R, we use the same noise bound parameter $\beta  = {\bar c^2}{\sigma ^2}$ such that $\mathbb{P}\left( {{{\left\| {{{\mathbf{b}}_i} - {\mathbf{R}}{{\mathbf{a}}_i}} \right\|}^2} \leq {{\bar c}^2}{\sigma ^2}} \right) = 1 - {10^{-6}}$ holds for inliers, which under Gaussian inlier noise can be computed from the 3-DoF Chi-squared distribution. For the rotation registration problem, AM alternatingly solves the two subproblems (\ref{eq_am_s}) and (\ref{eq27}), whilst AM-R alternatingly solves the two subproblems (\ref{eq15}) and (\ref{eq27}). We use the RANSAC solution as the initialization of AM and AM-R. All the runtime results of AM and AM-R include the runtime of the RANSAC initialization.

The Bunny dataset from the Stanford 3D Scanning Repository \cite{53} is used. Firstly, the point cloud is resized into a unit cube ${[0,1]^3}$ and randomly down-sampled to $N$ points with $N \in \{100,500\}$. Then, a random rotation is applied and additive noise and outliers are randomly generated according to (\ref{eq24}). Two conditions with low and high inlier noise are considered, with $\sigma = 0.01$ and $\sigma = 0.1$, respectively. Meanwhile, different outlier ratios from 0 to 95\% are considered. Each result is an average of 50 independent runs.

Fig. \ref{figure8} presents the rotation error of the algorithms in the two noise conditions for $N \!=\! 100$ and $N \!=\! 500$, respectively. It can be seen that FGR performs well at low outlier ratios, but tends to beak at relatively high outlier ratios, e.g., when the outlier ratio exceeds 70\% in the case of $N = 100$ and $\sigma = 0.1$. GORE-RANSAC generally has better performance than RANSAC as it firstly removes most of the outliers by the GORE method. For $N \!=\! 100$, QUASAR, AM and AM-R generally perform comparably and outperform the others. In the case of $N \!= \!500$, QUASAR is not compared as it runs out of memory when $N \!>\! 150$. For $N \!=\! 500$, SIME distinctly outperforms its counterparts in most cases, and the advantage is especially conspicuous in the high noise case.

Fig. \ref{figure9} compares the runtime of the algorithms. Clearly, FGR is the fastest. Though both QUASAR and AM-R involve solving SDP, AM-R is about 1000 times faster than QUASAR in the case of $N\!=\!100$. This thanks to the accelerating strategy using B-M factorization and the unconstrained formulation exploiting the sparsity of the problem. However, QUASAR cannot be accelerated like AM-R as it has a large number of constraints as explained in Remark 4.

\subsection{6-DoF Euclidean Registration}

\begin{figure*}[!t]
\centering
{{\includegraphics[width = 8.3cm]{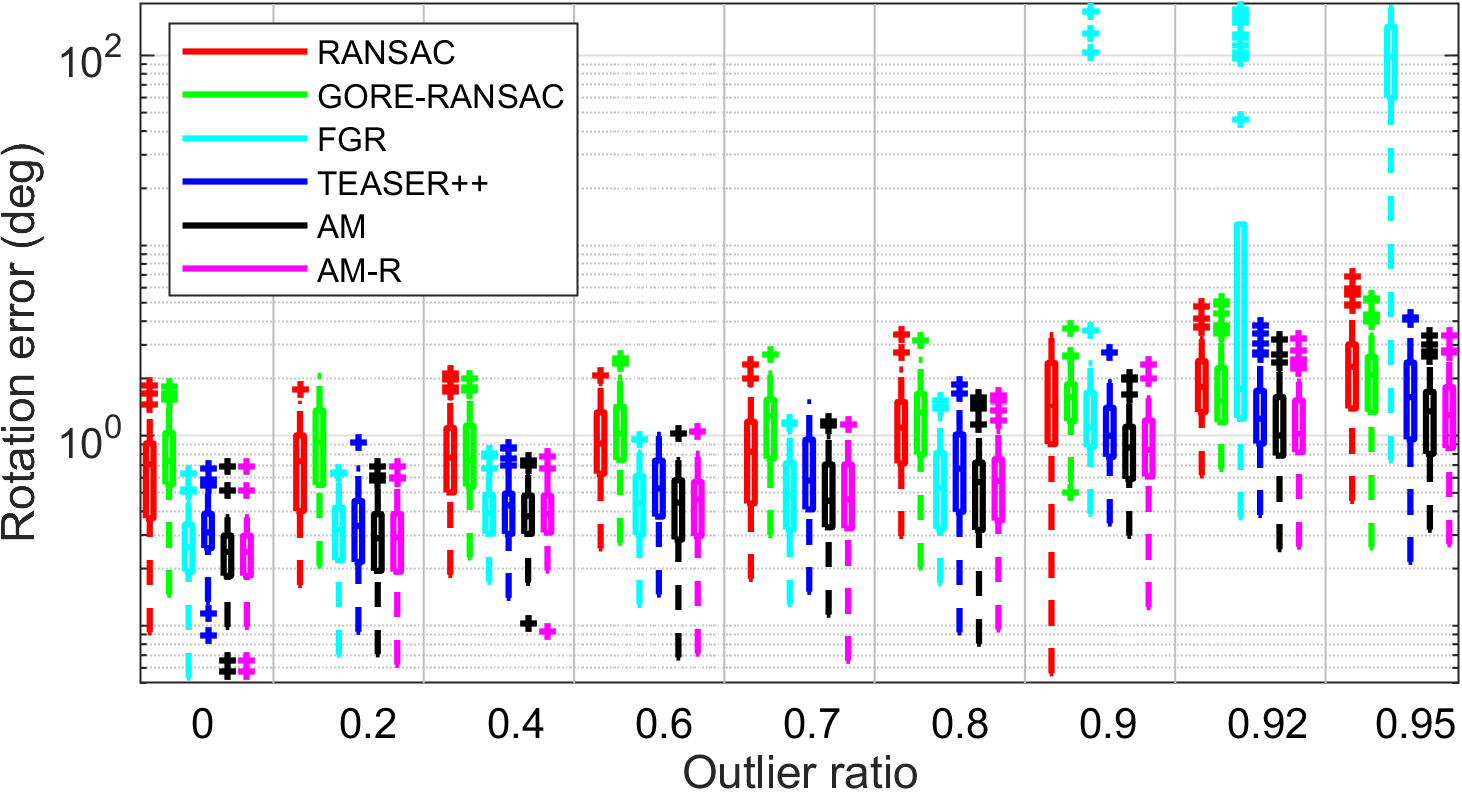}}}~~~~~~~
{{\includegraphics[width = 8.3cm]{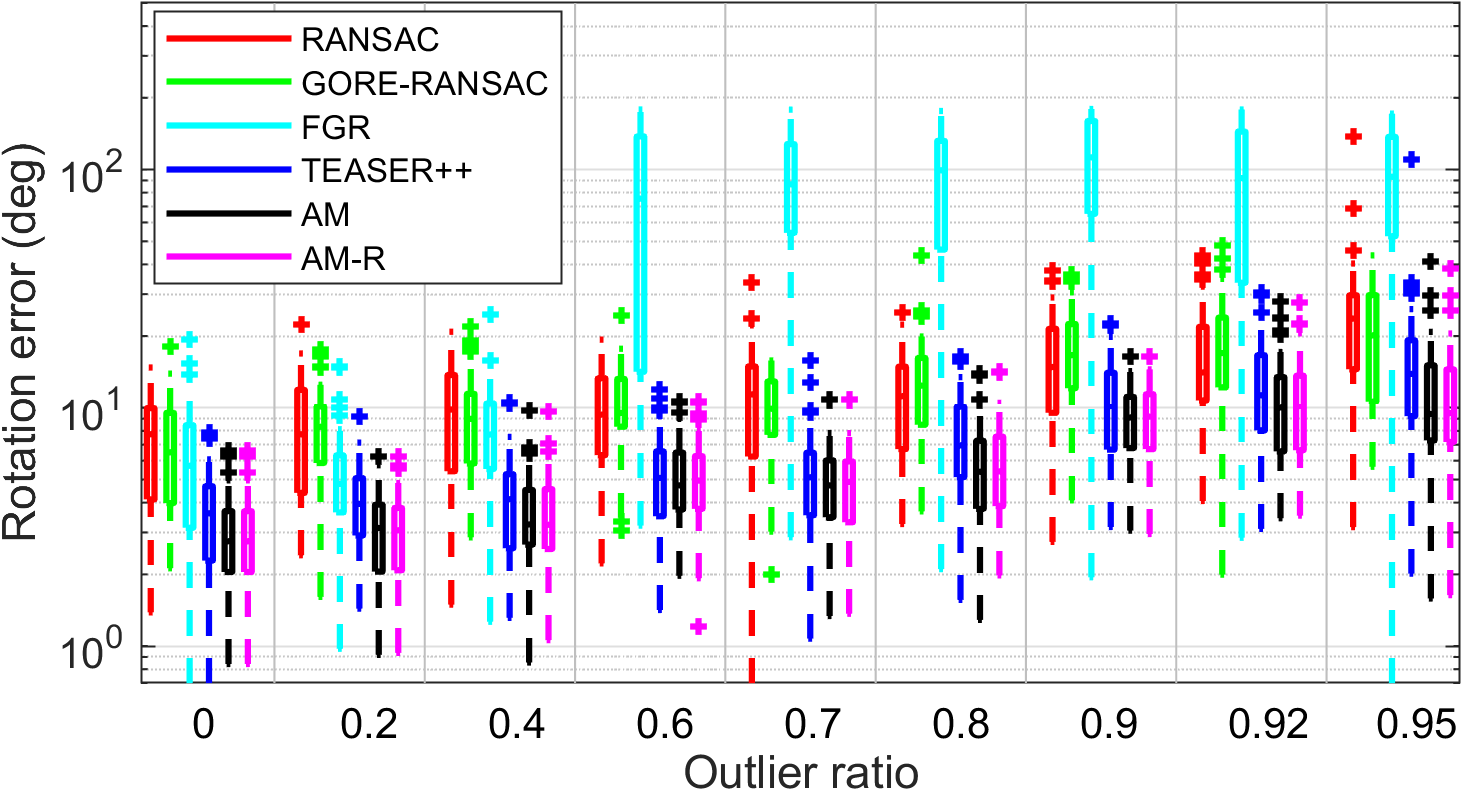}}}\\ \vspace{0.2cm}
{{\includegraphics[width = 8.3cm]{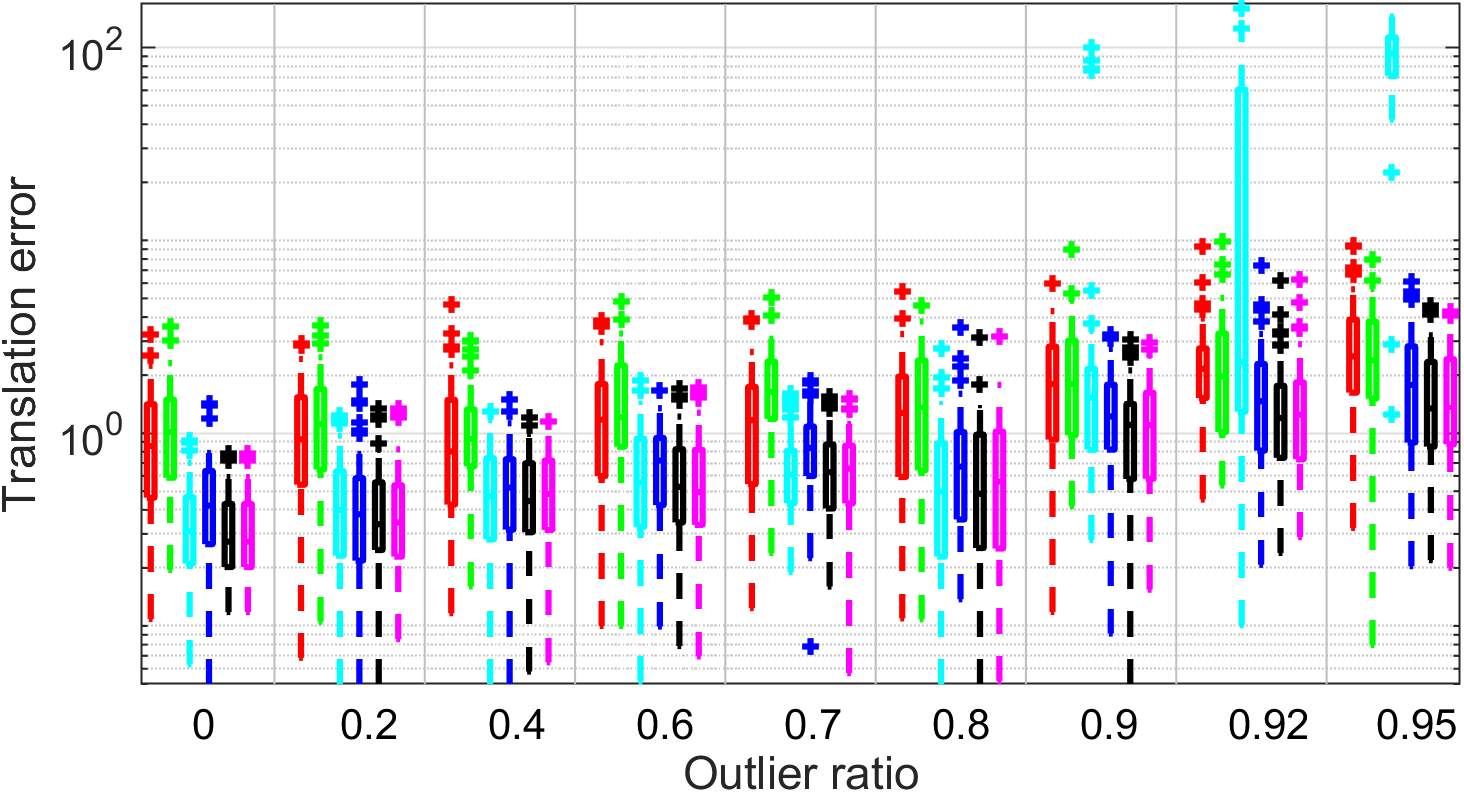}}}~~~~~~~
{{\includegraphics[width = 8.3cm]{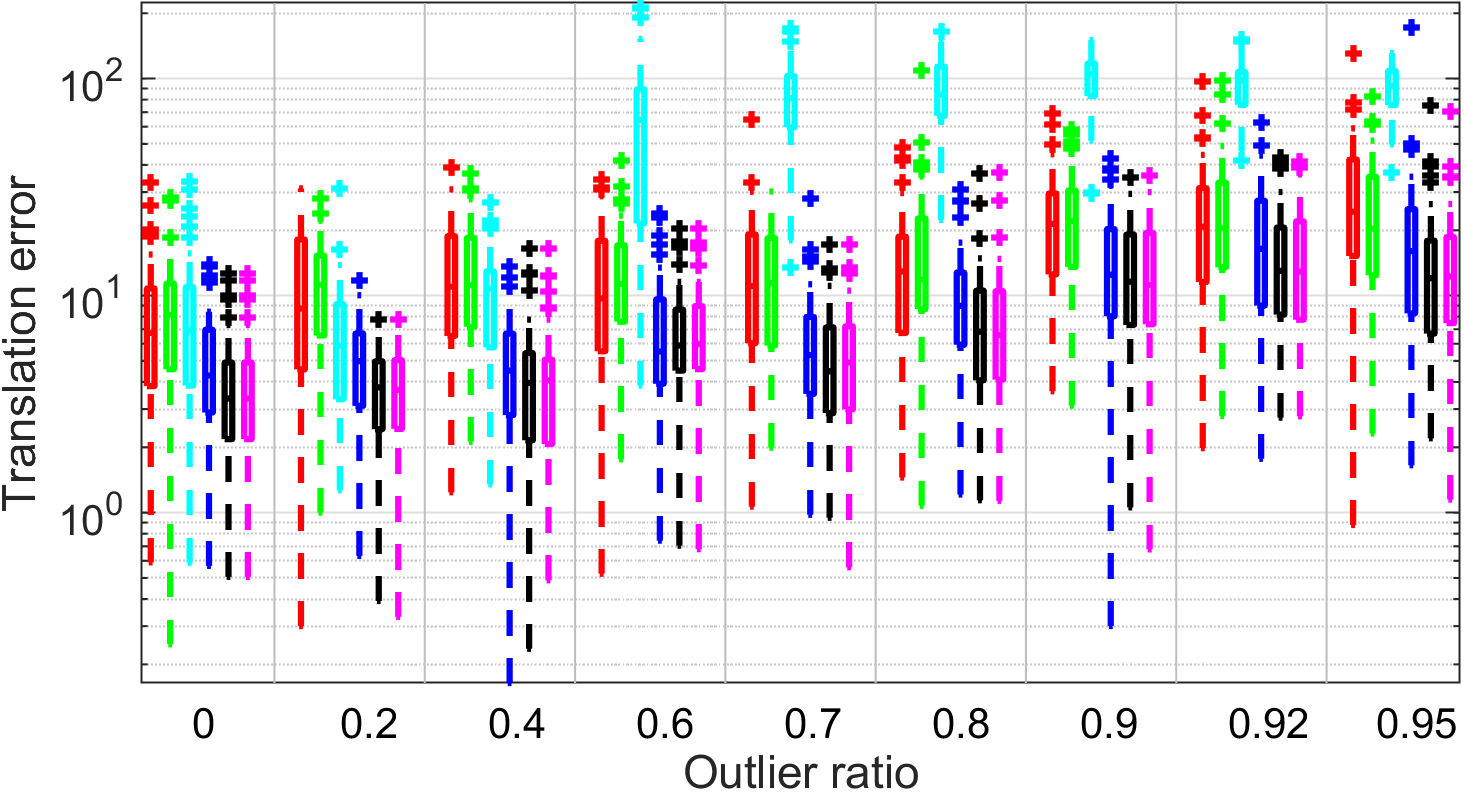}}}\\
\subfigure[$\sigma = 0.01$ (low noise)]{{\includegraphics[width = 8.3cm]{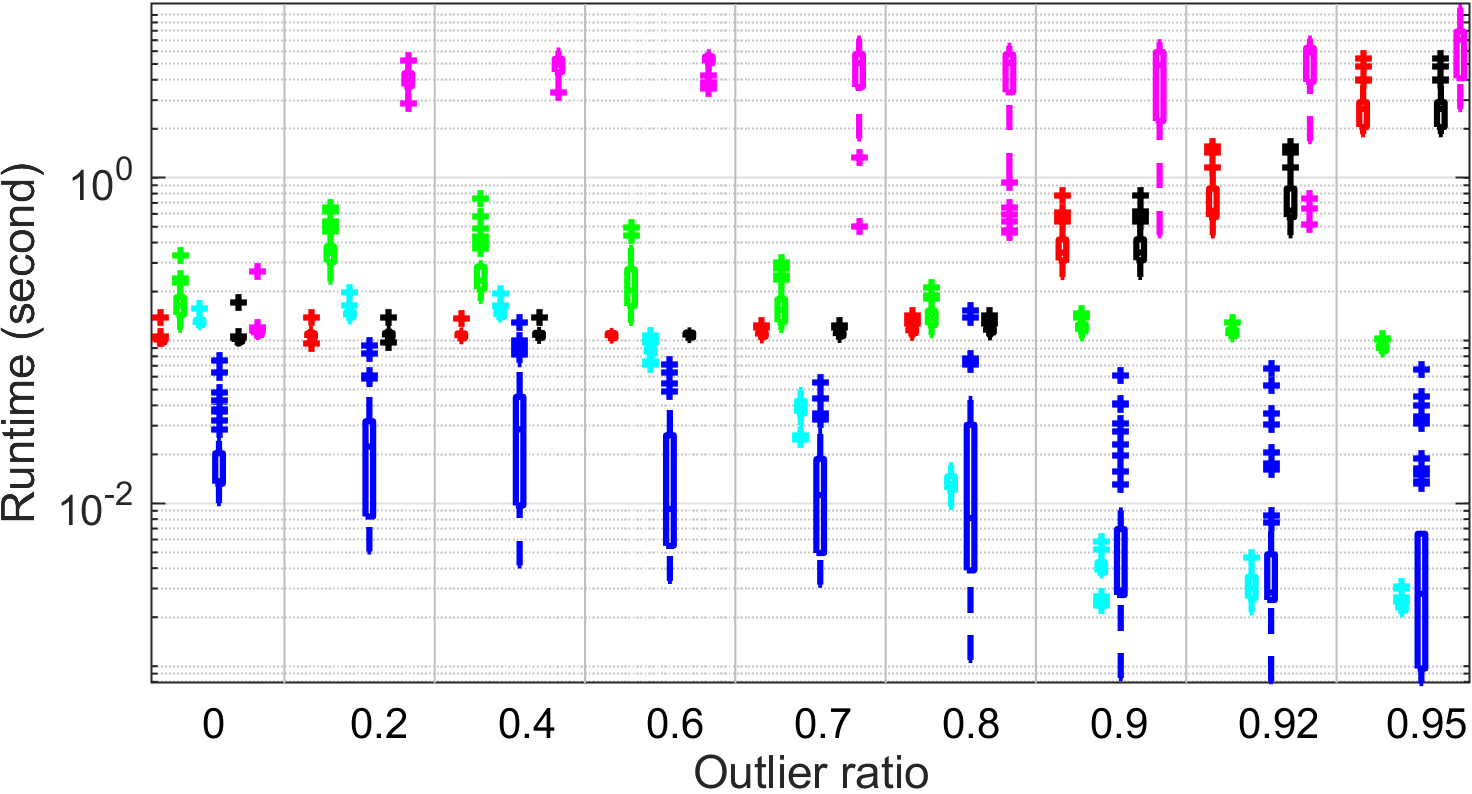}}}~~~~~~~
\subfigure[$\sigma = 0.1$ (high noise)]{{\includegraphics[width = 8.3cm]{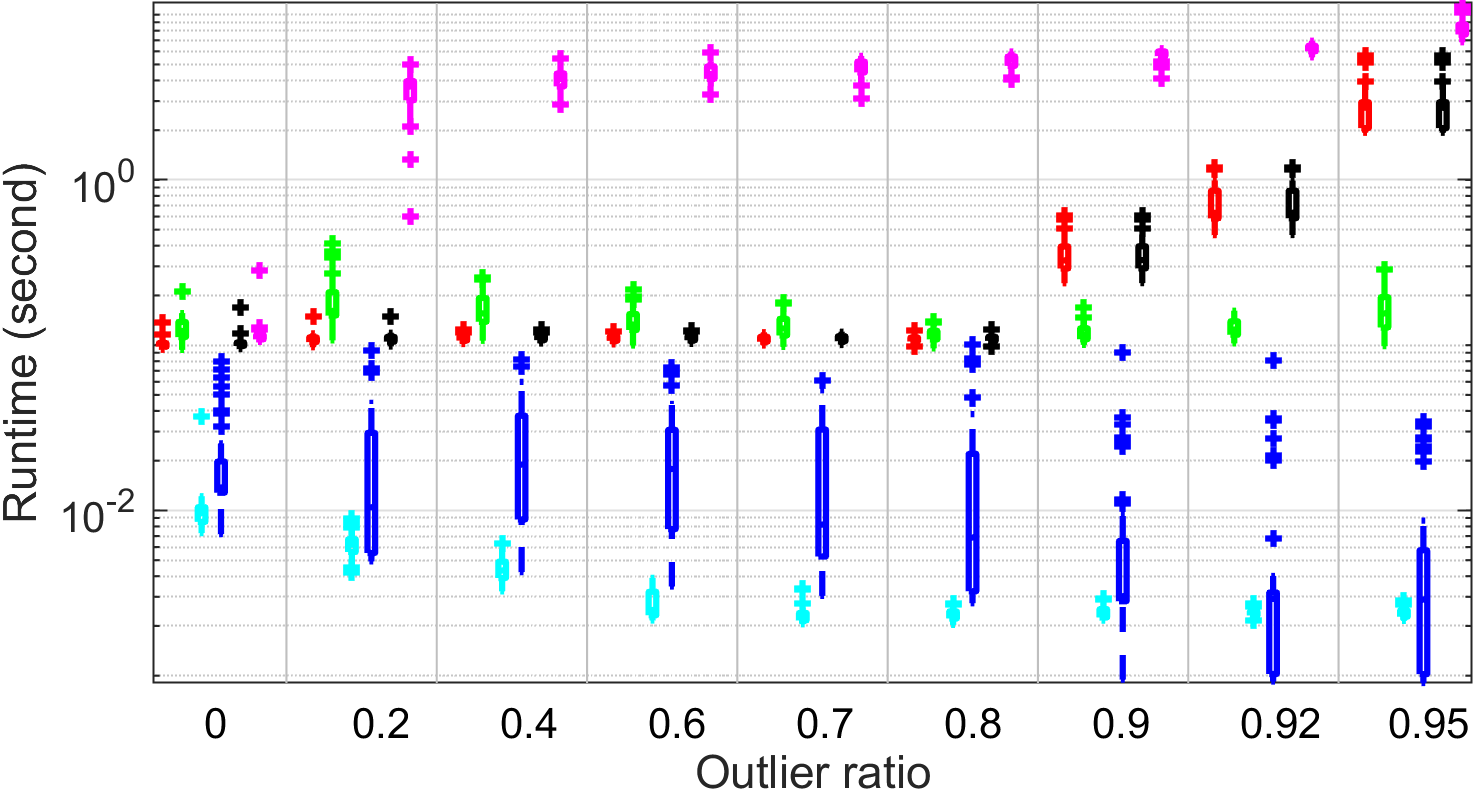}}}\\
\caption{Rotation error, translation error and runtime comparison in the 6-DoF Euclidean registration experiment. (a) $\sigma = 0.01$. (b) $\sigma = 0.1$.}
\label{figure10}
\end{figure*}

In the second experiment, we consider 6-DoF rigid registration, where both rotation and translation need to be estimated. We generate the data measurements similar to the rotation registration experiment except that a random translation is additionally considered according to (\ref{eq36}). In this setting, AM alternatingly solves the two subproblems (\ref{eq_am_s}) and (\ref{eq38}), whilst AM-R alternatingly solves the two subproblems (\ref{eq15}) and (\ref{eq38}). The TEASER++ algorithm (implemented in C++) \cite{51} is also compared in this experiment. TEASER++ decouples the scale, rotation and translation estimation and solves them separately and sequentially. It solves decoupled scale and translation estimation via adaptive voting, and solves the rotation estimation via a graduated nonconvexity scheme \cite{29}, which has shown highly effectiveness and efficiency. The noise bound parameter of TEASER++ is tuned to provide the best performance.

\begin{figure}[!t]
\centering
{\includegraphics[width = 8.7cm]{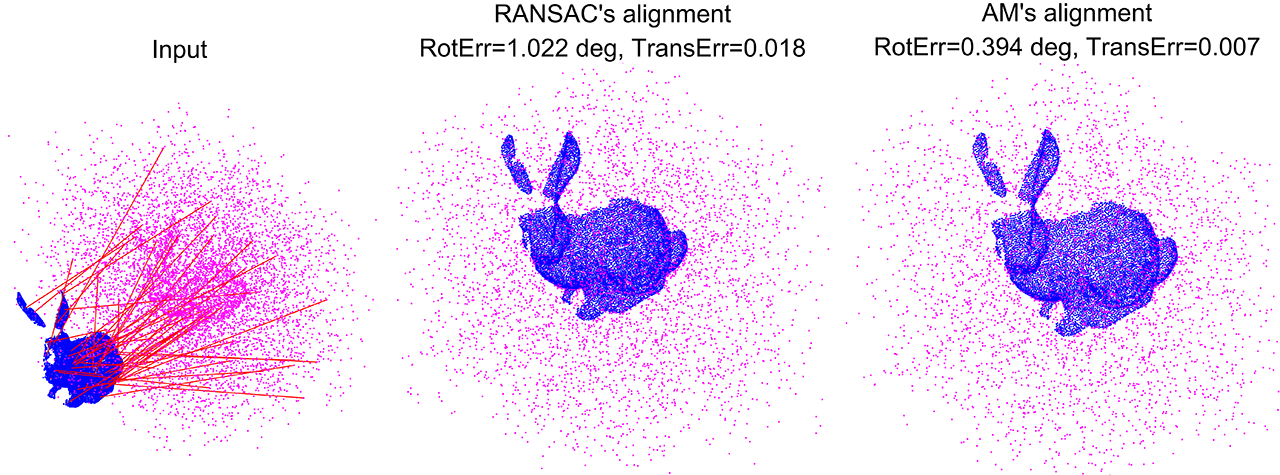}}
\caption{Illustration of 6-DoF Euclidean registration by RANSAC and the proposed AM algorithm (with outlier ratio being 80\%), along with the rotation error (RotErr) and translation error (TransErr).}
\label{figure11}
\end{figure}

Fig. \ref{figure10} presents the rotation error, translation error and runtime of the compared algorithms for $N=200$. Similar to the rotation registration experiment, two noise conditions with $\sigma = 0.01$ and $\sigma = 0.1$ are considered. It can be seen that FGR tends to break at high outlier ratios, especially in the high noise condition, e.g. when the outlier ratio exceeds 50\%. AM and AM-R achieve the best accuracy in most cases, and the advantage gets more prominent in the high noise condition. They again significantly outperform the RANSAC and GORE-RANSAC methods. Moreover, the results demonstrate the highly efficiency of FGR and TEASER++, which are much faster than AM-R. Fig. \ref{figure11} illustrates a typical registration example by RANSAC and AM at an outlier ratio of 80\%.

\subsection{Truncated $\ell_p$ Loss Under Non-Gaussian Noise}

The above experiments considers the truncated LS loss.
This experiment further investigates the truncated $\ell_p$ loss with $1\!\leq\! p\!\leq\! 2$.
The inlier noise is generated as generalized Gaussian distribution (GGD) with zero-mean as
\begin{equation}
p(x;v,\alpha) = \frac{v}{2\alpha \Gamma (1/v)}\exp{\left(-\frac{|x|^v}{\alpha^v}\right)},
\end{equation}
where $v$ is the shape parameter, $\alpha$ is the scale parameter, $\Gamma$ is the gamma function.
GGD adapts to a large family of symmetric distributions, spanning from Laplace ($v=1$) to Gaussian ($v=2$).
When $v<2$, the noise is super-Gaussian.

\begin{figure*}[!t]
\centering
\subfigure[Gaussian noise, $\sigma = 0.01$]{{\includegraphics[width = 5.8cm]{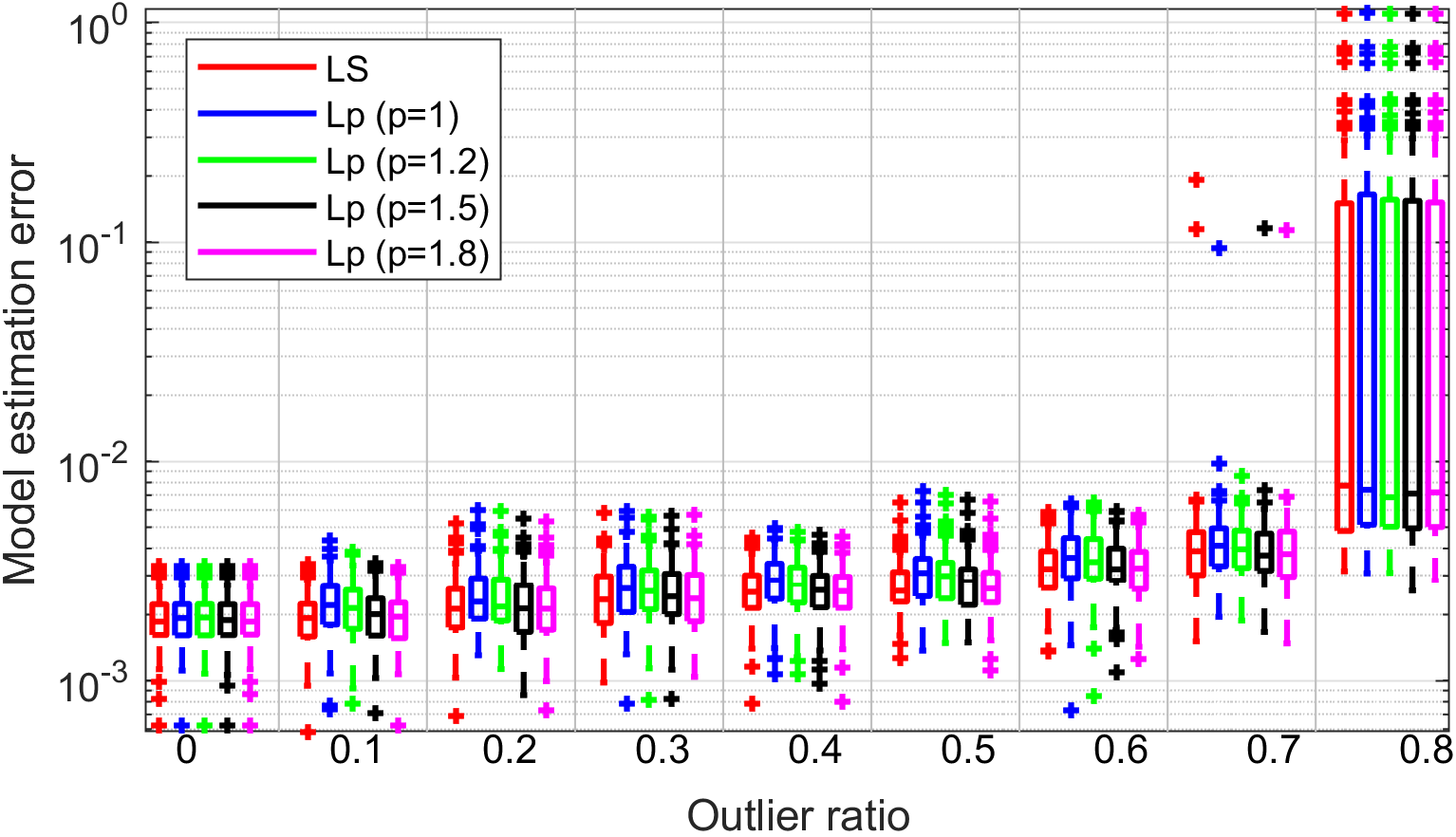}}}~
\subfigure[GGD noise with $v=0.5$, $\sigma = 0.01$]{{\includegraphics[width = 5.8cm]{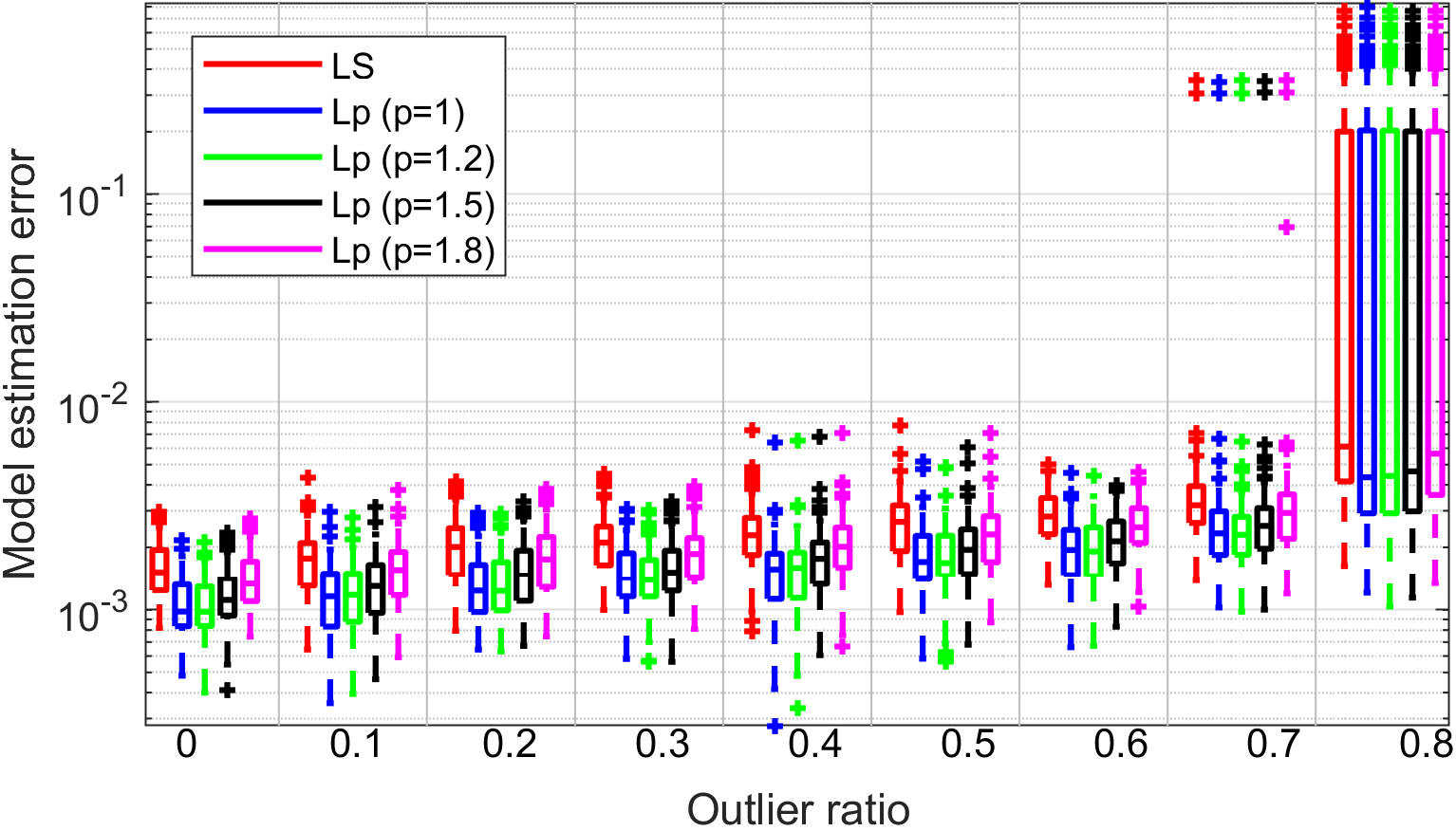}}}~
\subfigure[GGD noise with $v=0.5$, $\sigma = 0.1$]{{\includegraphics[width = 5.8cm]{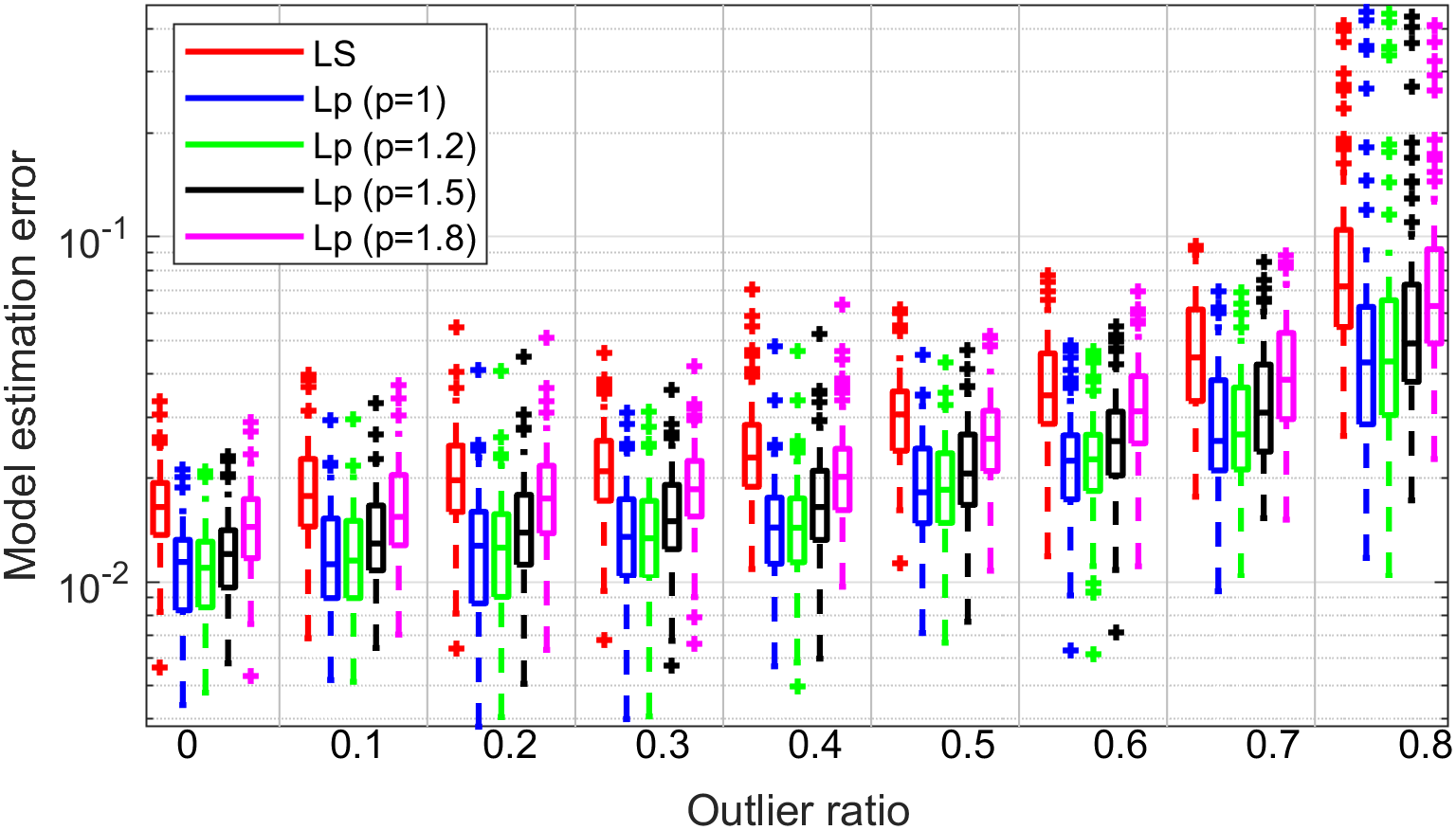}}}\\
\caption{Performance of the AM algorithm with truncated LS or $\ell_p$ loss in the linear regression experiment in the case of GGD inlier noise (with shape parameter $v$ and standard deviation $\sigma$) and uniformly distributed outliers in [-2, 2].}
\label{figure12}
\end{figure*}

\begin{figure*}[!t]
\centering
\subfigure[Gaussian noise, $\sigma = 0.01$]{{\includegraphics[width = 5.8cm]{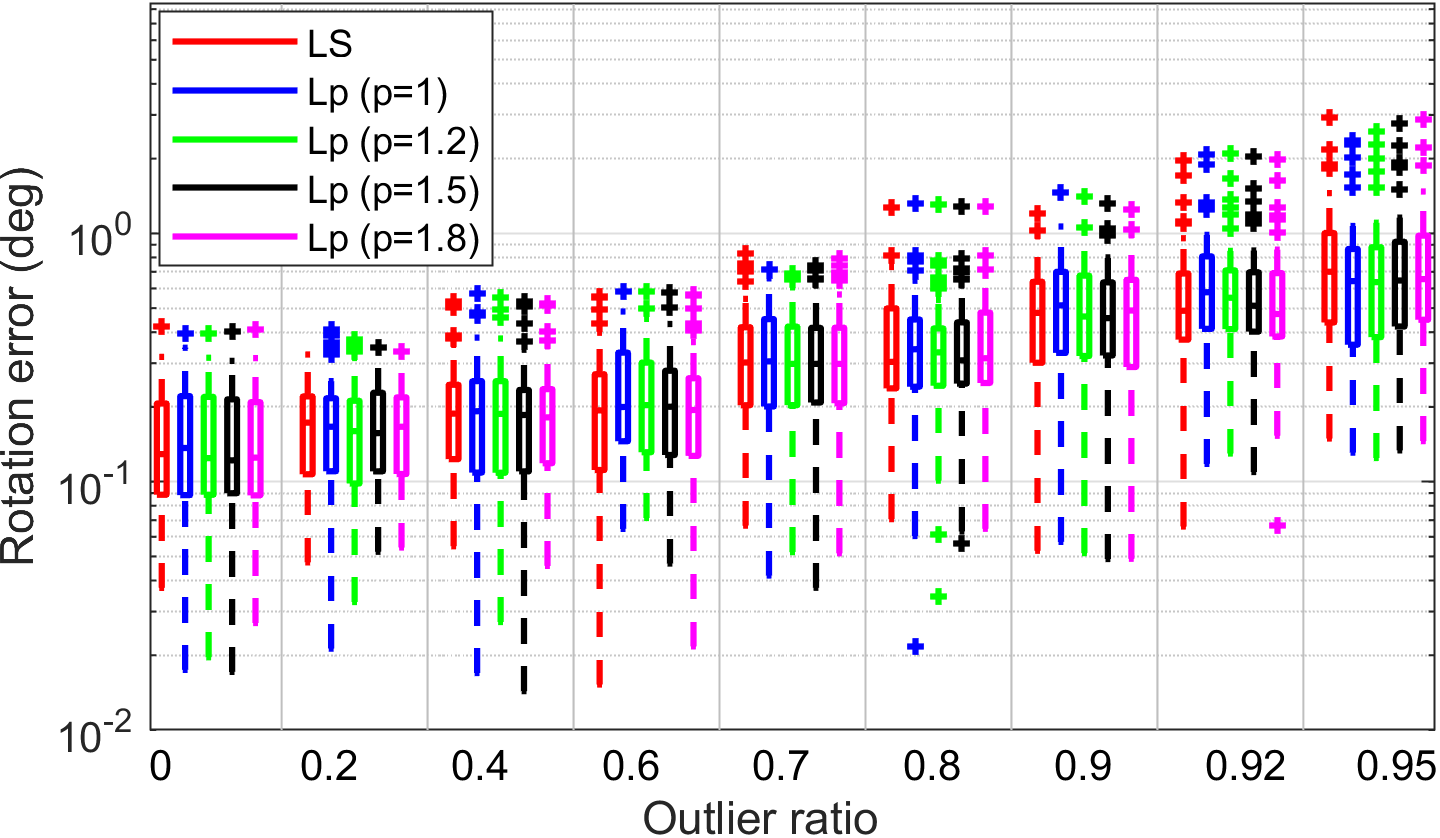}}}~
\subfigure[GGD noise with $v=0.5$, $\sigma = 0.01$]{{\includegraphics[width = 5.8cm]{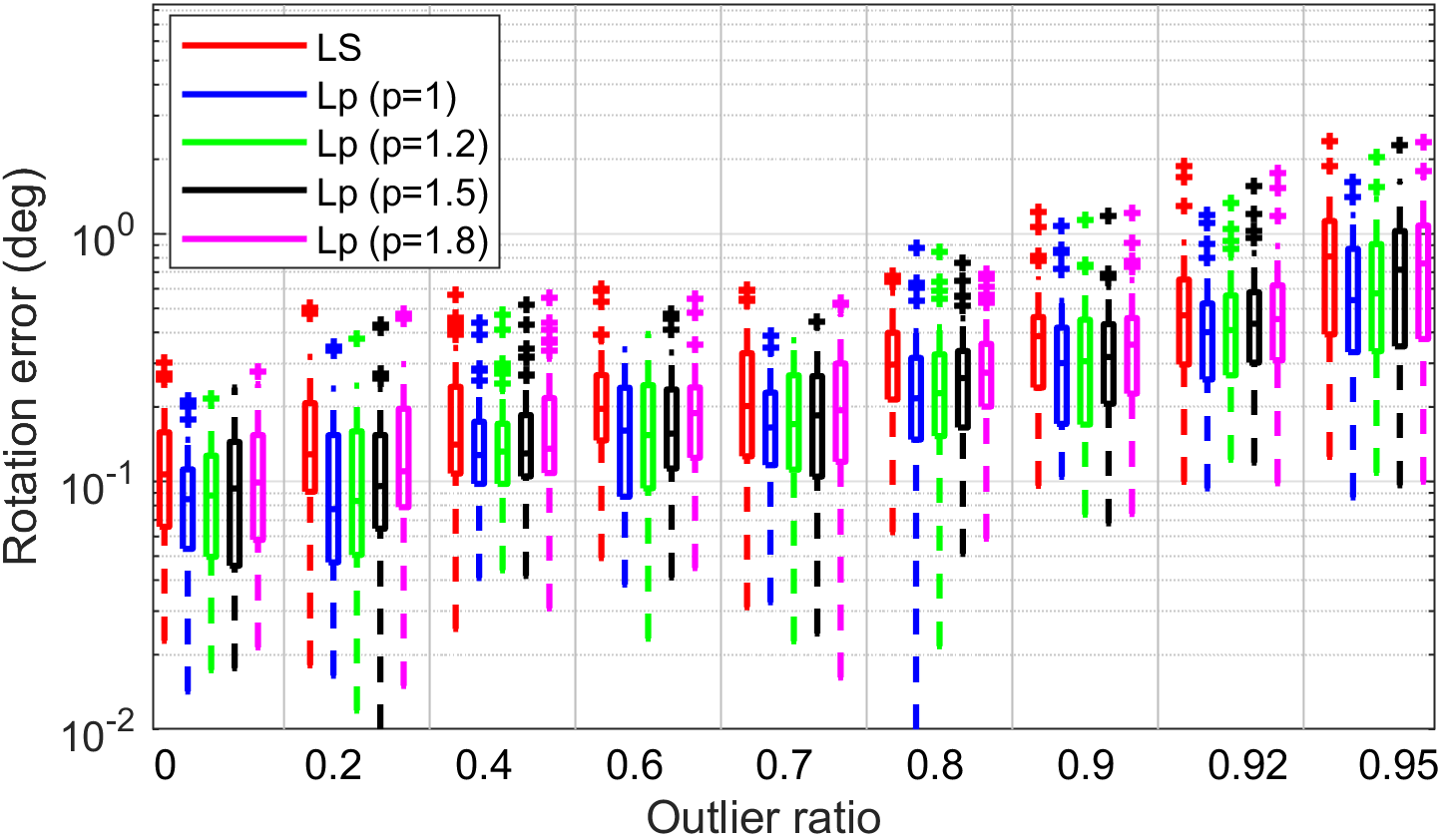}}}~
\subfigure[GGD noise with $v=0.5$, $\sigma = 0.1$]{{\includegraphics[width = 5.8cm]{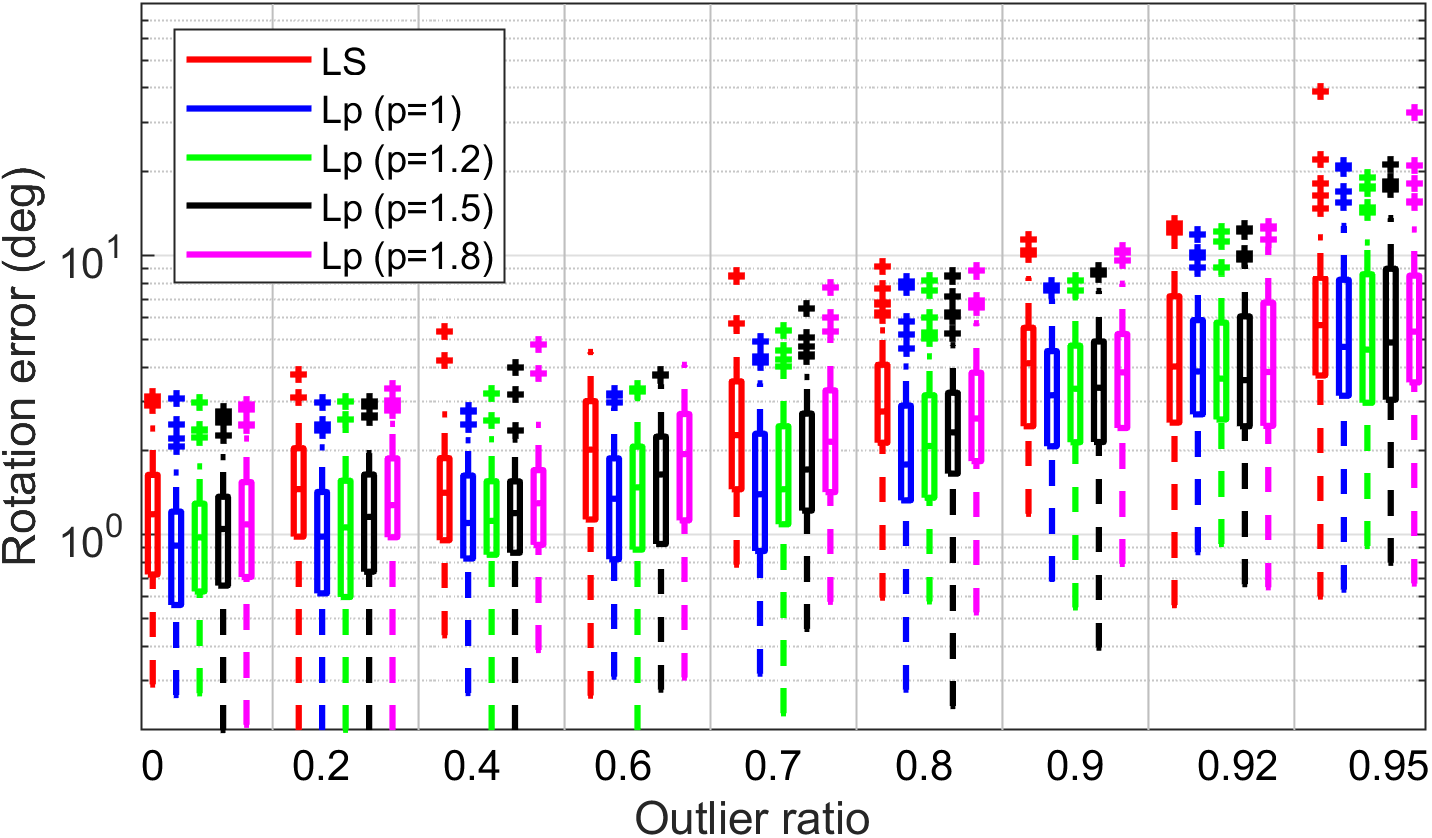}}}\\
\caption{Performance of the AM algorithm with truncated LS or $\ell_p$ loss in the rotation registration experiment (with $N=100$) in the case of GGD inlier noise (with shape parameter $v$ and standard deviation $\sigma$).}
\label{figure13}
\end{figure*}

Fig. \ref{figure12} shows the performance of the AM algorithm using the truncated LS and truncated $\ell_p$ loss in the linear regression experiment,
whilst Fig. \ref{figure13} shows that in the rotation registration experiment with $N=100$.
Two cases with Gaussian and GGD ($v=0.5$) inlier noise are considered.
It can be seen that in the case of GGD noise with $v=0.5$,
the truncated $\ell_p$ loss with a relatively small value of $p$ (e.g. $p=1$) can yield significant better accuracy than the truncated LS loss.
A more detailed comparison in various noise conditions is provided in Appendix \ref{Lp loss}.

\subsection{Real-world 3D Registration Experiment}
\label{sec:scan_matching}

We further evaluate the proposed SIME (AM) on a real-world scan matching task using the 3DMatch benchmark \cite{62}, which contains real indoor RGB-D scans from eight test scenes. For each scan, there are 5000 keypoints, on which putative correspondences are generated by nearest-neighbor matching on 3DSmoothNet descriptors \cite{63}. We compare the proposed SIME (AM) with RANSAC and TEASER++ on the same correspondence sets.
A registration result is counted as successful if the estimated rigid transformation has a rotation error smaller than $10^\circ$ and a translation error smaller than $30$ cm. For RANSAC, we set the maximum number of iterations to 10,000. For SIME (AM), we use the RANSAC estimate as initialization.

Table \ref{tab:scan_matching_3dmatch} reports the percentage of correctly registered scan pairs for each test scene, together with the average runtime per scan pair. The runtime only includes the robust transformation estimation stage, since feature extraction and descriptor matching are identical for all compared methods.
As can be seen, SIME (AM) achieves the best overall performance on the 3DMatch scan matching benchmark. Compared with TEASER++, SIME (AM) improves the mean registration success rate by 0.3\%, while compared with RANSAC the gain reaches 4.3\%. This result is consistent with our theoretical analysis and synthetic experiments that by jointly selecting inliers and refining the model under a truncated-loss objective, SIME is able to obtain a more reliable consensus set and a lower fitting residual than methods based purely on consensus maximization.

\begin{table*}[!t]
\centering
\small
\caption{Percentage of correct registration results on the 3DMatch scan matching benchmark (\%).}
\label{tab:scan_matching_3dmatch}
\setlength{\tabcolsep}{5pt}
\begin{tabular}{lcccccccccc}
\toprule
Method & Kitchen & Home 1 & Home 2 & Hotel 1 & Hotel 2 & Hotel 3 & Study & MIT Lab & Mean & Avg. Runtime (ms) \\
\midrule
RANSAC       & 96.2 & 91.7 & 74.5 & 91.6 & 84.6 & 90.7 & 82.2 & 81.8 & 86.7& 101.3 \\
TEASER++     & 97.8 & 92.3 & 82.7 & 96.9 & 88.5 & 94.4 & 88.7 & 84.4 & 90.7& 87.4 \\
SIME (AM)    & \textbf{98.2} & {92.3} & \textbf{83.2} & \textbf{97.3} & 88.5 & 94.4 & \textbf{89.4} & 84.4 &\textbf{91.0}& 112.9 \\
\bottomrule
\end{tabular}
\end{table*}

\section{Conclusion}
This paper investigated a truncated-loss based formulation for robust 3D registration using simultaneous inlier identification and model estimation (SIME). Compared with maximum consensus methods that only rely on hard inlier counting, SIME enables lower achievable fitting residuals in robust model fitting. To solve the resulting nonconvex problem, we developed two alternating minimization methods, namely a direct AM algorithm and an SDR-enhanced AM-R algorithm. We instantiated the proposed framework for both rotation registration and 6-DoF Euclidean registration. Experimental results on synthetic and real-world data demonstrated that the proposed methods achieve strong robustness under high noise and high outlier ratios, and compare favorably with representative robust registration methods. 


\appendices
\section{Proof of Proposition 1}
\label{proof:propo1}
\begin{proof}
Let $P\in(0,1]$ and $1-P\in[0,1)$ denote the prior probabilities of inliers and outliers, respectively. Then, under conditions i) and ii), the mixture density of $r_i$ is
\[
p(r_i\mid {\boldsymbol{\theta}})
=
Pc_i e^{-\Phi(r_i,\sigma_i)}
+
(1-P)\frac{1}{u-\tau}\mathbb{I}(\tau<r_i\le u).
\]
Thus, under conditions i) and ii), the likelihood of ${\mathbf{r}} = {[{r_1},{r_2}, \cdots ,{r_N}]^T}$ is given by
\[\begin{gathered}
\mathcal{L}({\mathbf{r}};{\boldsymbol{\theta}}) = \prod\limits_{i = 1}^N {{{\left[ {P\exp \left( { - \Phi ({r_i},{\sigma _i})} \right)} \right]}^{1 - \mathbb{I}(\tau  < {r_i} \leq u)}}}  \hfill \\
~~~~~~~~~~~~~~~~~~~~ \times {\left[ {P\exp \left( { - \Phi ({r_i},{\sigma _i})} \right) + \frac{{1 - P}}{{u - \tau }}} \right]^{\mathbb{I}(\tau  < {r_i} \leq u)}} \hfill, \\
\end{gathered} \]
with $h({\boldsymbol{\theta}})=0$. The corresponding log-likelihood is
\begin{equation} \label{eq06}
\begin{split}
&\log \mathcal{L}({\mathbf{r}};{\boldsymbol{\theta}}) = \sum\limits_{i = 1}^N \left({1 - \mathbb{I}(\tau  < {r_i} \leq u)}\right) \left[{ - \Phi ({r_i},{\sigma _i})} + \log P\right]\\
&~~~~ + {\mathbb{I}(\tau  < {r_i} \leq u)} \mathop {\underbrace{\log \left[ {P\exp \left( { - \Phi ({r_i},{\sigma _i})} \right) + \frac{{1 - P}}{{u - \tau }}} \right]}}\limits_{{L_o}({r_i})},
\end{split}
\end{equation}
with $h({\boldsymbol{\theta}})=0$. Since $\Phi $ is an increasing function, for ${r_i} \in (\tau ,u]$ it follows that
\[\log (1 - P) - \log (u - \tau ) < {L_o}({r_i}) < {L_o}(\tau ).\]
That is as ${r_i}$ increases on $(\tau ,u]$, ${L_o}({r_i})$ decreases from ${L_o}(\tau )$ toward $\log (1 - P) - \log (u - \tau )$.
Meanwhile, for sufficiently large $u$, we have $\exp \left( { - \Phi ({r_i},{\sigma _i})} \right) \to 0$ for ${r_i} > u$.
Hence, the likelihood is dominated by the two cases of ${r_i} \le \tau $ and $\tau  < {r_i} \le u$, i.e., ${r_i} \le u$.
Then, from the fact that $\mathbb{I}(\tau  < {r_i} \leq u)$ is equivalent to $\mathbb{I}(\Phi (\tau ) < \Phi ({r_i}) \leq \Phi (u))$,
it is easy to see that with some $\beta $ satisfying $\log (1 - P) - \log (u - \tau ) \leq \beta  \leq {L_o}(\tau )$, the minimization problem
\[\mathop {{\text{min }}}\limits_{{\boldsymbol{\theta}} \in {\mathbb{R}^d},{s_i} \in \{ 0,1\} } \sum\nolimits_{i = 1}^N {(1 - {s_i})\Phi ({r_i},{\sigma _i}) + \beta {s_i}}, \]
can be viewed as an approximation of maximizing the log-likelihood (\ref{eq06}).
Under i.i.d assumption of inliers with ${\sigma _i} = \sigma $ for $\forall i \in {I^*}$, taking the scale parameter $\sigma $ into $\beta$, and ignoring constant terms independent on ${r_i}$, it finally leads to (\ref{eq_sime}).
\end{proof}

\section{Proof of Proposition 2}
\label{proof:propo2}

\begin{proof}
Let $({\boldsymbol{\theta}}^+,{\mathbf{s}}^+)$ be a global solution of SIME, and let
\[
I^+:=\{i:s_i^+=0\}
\]
be its inlier set. Since, for fixed ${\boldsymbol{\theta}}^+$, each binary variable $s_i^+$ minimizes
\[
(1-s_i)\Phi(r_i({\boldsymbol{\theta}}^+))+\Phi(\tau)s_i
\quad \text{over } s_i\in\{0,1\},
\]
it follows that
\[
\Phi(r_i({\boldsymbol{\theta}}^+))\le \Phi(\tau),\qquad \forall i\in I^+.
\]
Because $\Phi$ is increasing on $[0,+\infty)$, we have
\[
r_i({\boldsymbol{\theta}}^+)\le \tau,\qquad \forall i\in I^+.
\]
Hence $I^+$ is a consensus set. Moreover, since $I^*$ is a maximum consensus set, it follows that
\[
|I^+|\le |I^*|.
\]

Now assume that $\Phi(r_i({\boldsymbol{\theta}}_{I^*}))\le \Phi(\tau)$ for all $i\in I^*$. Then, by monotonicity of $\Phi$,
\[
r_i({\boldsymbol{\theta}}_{I^*})\le \tau,\qquad \forall i\in I^*.
\]
Moreover, since $I^*$ is a maximum consensus set, there cannot exist any $j\in \Omega\backslash I^*$ such that
\[
r_j({\boldsymbol{\theta}}_{I^*})\le \tau,
\]
for otherwise $I^*\cup\{j\}$ would be a larger consensus set, contradicting the maximality of $I^*$. Therefore,
\[
r_j({\boldsymbol{\theta}}_{I^*})>\tau,\qquad \forall j\in \Omega\backslash I^*,
\]
and hence
\[
\Phi(r_j({\boldsymbol{\theta}}_{I^*}))>\Phi(\tau),\qquad \forall j\in \Omega\backslash I^*.
\]
Thus we have $I^*\in\digamma$, and
\[
f({\boldsymbol{\theta}}_{I^*},{\mathbf{s}}_{I^*})
=
R(I^*)+|\Omega\backslash I^*|\Phi(\tau).
\]

For any other consensus set $I^\bullet\in\digamma$, we similarly have
\[
f({\boldsymbol{\theta}}_{I^\bullet},{\mathbf{s}}_{I^\bullet})
=
R(I^\bullet)+|\Omega\backslash I^\bullet|\Phi(\tau).
\]
Condition (\ref{eq07}) is equivalent to
\[
R(I^\bullet)+|\Omega\backslash I^\bullet|\Phi(\tau)
>
R(I^*)+|\Omega\backslash I^*|\Phi(\tau),
\]
that is,
\[
f({\boldsymbol{\theta}}_{I^\bullet},{\mathbf{s}}_{I^\bullet})
>
f({\boldsymbol{\theta}}_{I^*},{\mathbf{s}}_{I^*}),
\qquad \forall I^\bullet\in\digamma,\ I^\bullet\neq I^*.
\]
Therefore, $({\boldsymbol{\theta}}_{I^*},{\mathbf{s}}_{I^*})$ attains the minimum objective value of SIME, and the corresponding inlier set is $I^*$. Hence $I^+=I^*$.
\end{proof}

\section{Proof of Theorem 3}\label{proof:thrm3}
\begin{proof}
Let $({\boldsymbol{\theta}}^+,{\mathbf{s}}^+)$ be a global solution of SIME, and let
\[
I^+:=\{i:s_i^+=0\}
\]
be its inlier set. By Proposition 2, $I^+$ is a consensus set, and hence
\[
|I^+|\le |I^*|.
\]

Moreover, since $({\boldsymbol{\theta}}^+,{\mathbf{s}}^+)$ is globally optimal for SIME, ${\boldsymbol{\theta}}^+$ must minimize
\[
\sum_{i\in I^+}\Phi(r_i({\boldsymbol{\theta}}))
\]
subject to $h({\boldsymbol{\theta}})=0$ for the fixed support $I^+$. Therefore,
\[
{\boldsymbol{\theta}}^+={\boldsymbol{\theta}}_{I^+},
\]
and
\[
f({\boldsymbol{\theta}}^+,{\mathbf{s}}^+)
=
R(I^+)+|\Omega\backslash I^+|\Phi(\tau).
\]
On the other hand, by definition of ${\boldsymbol{\theta}}_{I^*}$ and ${\mathbf{s}}_{I^*}$, we have
\[
f({\boldsymbol{\theta}}_{I^*},{\mathbf{s}}_{I^*})
=
R(I^*)+|\Omega\backslash I^*|\Phi(\tau).
\]

Since $({\boldsymbol{\theta}}^+,{\mathbf{s}}^+)$ is a global solution of SIME, we have
\[
f({\boldsymbol{\theta}}^+,{\mathbf{s}}^+)
\le
f({\boldsymbol{\theta}}_{I^*},{\mathbf{s}}_{I^*}),
\]
that is,
\[
R(I^+)+|\Omega\backslash I^+|\Phi(\tau)
\le
R(I^*)+|\Omega\backslash I^*|\Phi(\tau).
\]
Using $|I^+|\le |I^*|$, equivalently $|\Omega\backslash I^+|\ge |\Omega\backslash I^*|$, it follows that
\[
R(I^+)
\le
R(I^*)-\bigl(|I^*|-|I^+|\bigr)\Phi(\tau)
\le
R(I^*).
\]
By the definitions of $R(I^+)$ and $R(I^*)$, this proves (\ref{eq10}).

Next, suppose that $I^*\notin\digamma$. Then, by definition of $\digamma$, the binary vector ${\mathbf{s}}_{I^*}$ is not an optimal choice for the fixed parameter ${\boldsymbol{\theta}}_{I^*}$. Hence there exists some $\tilde{\mathbf{s}}\in\{0,1\}^N$ such that
\[
f({\boldsymbol{\theta}}_{I^*},\tilde{\mathbf{s}})
<
f({\boldsymbol{\theta}}_{I^*},{\mathbf{s}}_{I^*}).
\]
Since $({\boldsymbol{\theta}}^+,{\mathbf{s}}^+)$ is globally optimal for SIME, we further have
\[
f({\boldsymbol{\theta}}^+,{\mathbf{s}}^+)
\le
f({\boldsymbol{\theta}}_{I^*},\tilde{\mathbf{s}})
<
f({\boldsymbol{\theta}}_{I^*},{\mathbf{s}}_{I^*}).
\]
Therefore,
\[
R(I^+)+|\Omega\backslash I^+|\Phi(\tau)
<
R(I^*)+|\Omega\backslash I^*|\Phi(\tau).
\]
With the use of $|\Omega\backslash I^+|\ge |\Omega\backslash I^*|$, we have
\[
R(I^+)
<
R(I^*)-\bigl(|I^*|-|I^+|\bigr)\Phi(\tau)
\le
R(I^*).
\]
Hence the inequality in (\ref{eq10}) is strict.
\end{proof}

\section{Derivation of (\ref{eq13})}
\label{deriv_sdr}

To show the equivalence between the SIME formulation (\ref{eq_sime}) and the formulation (\ref{eq13}), it is enough to show the equivalence between
\begin{equation}\label{eq:d-1}
\min_{\bm{\theta}\in\mathbb{R}^d,\ \mathbf{s}\in\{0,1\}^N}
\sum_{i=1}^N (1-s_i)\Phi\bigl(r_i(\bm{\theta})\bigr)+\beta s_i,
\end{equation}
and
\begin{equation}\label{eq:d-2}
\begin{split}
&\mathop {{\text{min }}}\limits_{{\boldsymbol{\theta}} \in {\mathbb{R}^d},{\mathbf{S}} \in {\mathbb{S}^{(N + 1)}}} {\text{tr}}({\mathbf{\Lambda S}})\\
\mathrm{subject~to~~} &\operatorname{diag}(\mathbf{S})=\mathbf{1}_{N+1}, \mathbf{S}\succeq 0, \operatorname{rank}(\mathbf{S})=1,
\end{split}
\end{equation}
where
\[
\mathbf{\Lambda}=
\begin{bmatrix}
0 & (\beta-\bm{\Phi}^T)/2\\[2pt]
(\beta-\bm{\Phi})/2 & \operatorname{diag}(\bm{\Phi})
\end{bmatrix},
\]
with
\[
\bm{\Phi}=[\Phi_1,\Phi_2,\ldots,\Phi_N]^T,
\quad
\Phi_i:=\Phi\bigl(r_i(\bm{\theta})\bigr).
\]

First, problem (\ref{eq:d-1}) can be equivalently expressed as
\begin{equation}\label{eq:d-3}
\min_{\bm{\theta}\in\mathbb{R}^d, \mathbf{s}\in\{-1,1\}^N}
\frac{1}{2}\sum_{i=1}^N (1-s_i)\Phi\bigl(r_i(\bm{\theta})\bigr)+(1+s_i)\beta,
\end{equation}
where we changed $\mathbf{s}\in\{0,1\}^N$ to $\mathbf{s}\in\{-1,1\}^N$. 
Then, ignoring the constant terms, it is easy to see that problem (\ref{eq:d-3}) can be equivalently expressed as
\begin{equation}\label{eq:d-5}
\min_{\bm{\theta}\in\mathbb{R}^d, \mathbf{s}\in\{-1,1\}^N}
(\beta-{\bm{\Phi}})^T\mathbf{s}+\mathbf{s}^T\operatorname{diag}(\bm{\Phi})\mathbf{s}.
\end{equation}

Let
\[
\bar{\mathbf{s}}=[t,\mathbf{s}^T]^T\in\{-1,1\}^{N+1},
\]
then the objective of (\ref{eq:d-5}) can be rewritten as $\bar{\mathbf{s}}^T{\bm{\Lambda}} \bar{\mathbf{s}}$. Here, an auxiliary variable $t\in\{-1,1\}$ is introduced, with which $\mathbf{s}/t$ will be the solution. Meanwhile, define
\[
\mathbf{S}:=\bar{\mathbf{s}}\bar{\mathbf{s}}^T.
\]
It is easy to see that $\mathbf{S}=\bar{\mathbf{s}}\bar{\mathbf{s}}^T$ with constraint $\bar{\mathbf{s}}\in\{-1,1\}^{N+1}$ is equivalent to
\[
\operatorname{diag}({\mathbf{S}})=\mathbf{1}_{N+1},\quad \mathbf{S}\succeq 0,\quad \operatorname{rank}(\mathbf{S})=1.
\]
Thus, with $\bar{\mathbf{s}}^T{\bm{\Lambda}}\bar{\mathbf{s}}=\operatorname{tr}(\mathbf{\Lambda S})$, problem (\ref{eq:d-5}) can be rewritten as (\ref{eq:d-2}), which justifies the equivalence between (\ref{eq:d-1}) and (\ref{eq:d-2}), and finally leads to (\ref{eq13}).

\section{Truncated $\ell_p$ Loss}
\label{Lp loss}
Besides the truncated LS loss and truncated reprojection error, the SIME formulation can also adapt to other general truncated loss functions. Here we take the truncated $\ell_p$ (TLP) loss with $p\ge 1$ as an example. Obviously, TLP loss degenerates to the truncated $\ell_1$ loss and the truncated LS loss when $p=1$ and $p=2$, respectively. When the inlier noise is Gaussian distributed, $p=2$ is the optimal choice. When the inlier noise follows a super-Gaussian distribution, the optimal value of $p$ should be $p<2$, e.g. $p=1$ should be the optimal choice for Laplacian distribution. Generally, a smaller value of $p$ should be used when the inlier noise distribution has a thicker tail (more impulsive the inlier noise is). We restrict $p\ge 1$ as in this case $\Phi$ is convex. Meanwhile, since in the SIME formulation the outliers are modeled as a wide uniform distribution (see Proposition 1), it is reasonable to assume that the inlier noise is not too impulsive and hence we can restrict $p\ge 1$.

With the linear residual model as presented in Case~1 in Section~3-A, which is encountered in the problems including robust linear regression, homography estimation with algebraic distance, and affinity estimation, if $\Phi$ is chosen as the $\ell_p$ loss, the model fitting term in SIME becomes
\[
\Phi\bigl(r_i(\bm{\theta})\bigr)=|b_i-\mathbf{a}_i^T\bm{\theta}|^p.
\]
Then, the $\bm{\theta}$-subproblem (8) of SIME can be solved by the iteratively reweighted LS algorithm as
\begin{equation}
\bm{\theta}^{k+1}
=
\arg\min_{\bm{\theta}\in\mathbb{R}^d}
\sum_{i=1}^N (1-s_i)w_i^{k+1}(b_i-\mathbf{a}_i^T\bm{\theta})^2,
\label{A7}
\end{equation}
subject to $h(\bm{\theta})=0$, 
where
\[
w_i^k=\bigl|\mathbf{a}_i^T\bm{\theta}^k-b_i\bigr|^{p-2}.
\]
Problem (\ref{A7}) is quadratic and $\bm{\theta}^{k+1}$ can be computed in closed form.

Similarly, for the rotation registration problem, the $\bm{\theta}$-subproblem (24) can be solved by the iteratively reweighted LS algorithm as
\begin{equation}
\begin{split}
{\bm{\theta}}^{k+1}&=\arg\min_{{\bm{\theta}}\in\mathbb{R}^4}
\sum_{i=1}^N (1-S_{1,i+1})w_i^{k+1}\|\hat{\mathbf{b}}_i-\bm{\theta}\circ\hat{\mathbf{a}}_i\circ\bm{\theta}^{-1}\|^2\\
&\text{s.t.~~} \|{\bm{\theta}}\|=1,
\label{A8}
\end{split}
\end{equation}
where
\[
w_i^{k+1}=\|\hat{\mathbf{b}}_i-{\bm{\theta}}^{k}\circ\hat{\mathbf{a}}_i\circ({\bm{\theta}^{k}})^{-1}\|^{p-2}.
\]
Then, problem (\ref{A8}) can be solved in closed form via eigen-decomposition similar to (25). The 6-DoF Euclidean registration problem with $\ell_p$ loss can be solved by the reweighted LS algorithm similarly.

As the truncated LS loss and truncated reprojection error have been well investigated in Section~V, here we investigate the truncated $\ell_p$ loss with $1\le p<2$ in comparison with the truncated LS loss. The inlier noise is generated as generalized Gaussian distribution (GGD) with zero mean, i.e.,
\begin{equation}
p(x;v,\sigma)
=
\frac{v}{2\sigma\,\Gamma(1/v)}
\exp\!\left(-\frac{|x|^v}{\sigma^v}\right),
\tag{A9}
\end{equation}
where $v>0$ is the shape parameter, $\sigma>0$ is the scale parameter, and $\Gamma(\cdot)$ is the gamma function. GGD adapts to a large family of symmetric distributions, spanning from Laplace ($v=1$) to Gaussian ($v=2$). When $v<2$, the noise is super-Gaussian.

\begin{figure}[ht]
\centering
{\includegraphics[width = 8.3cm]{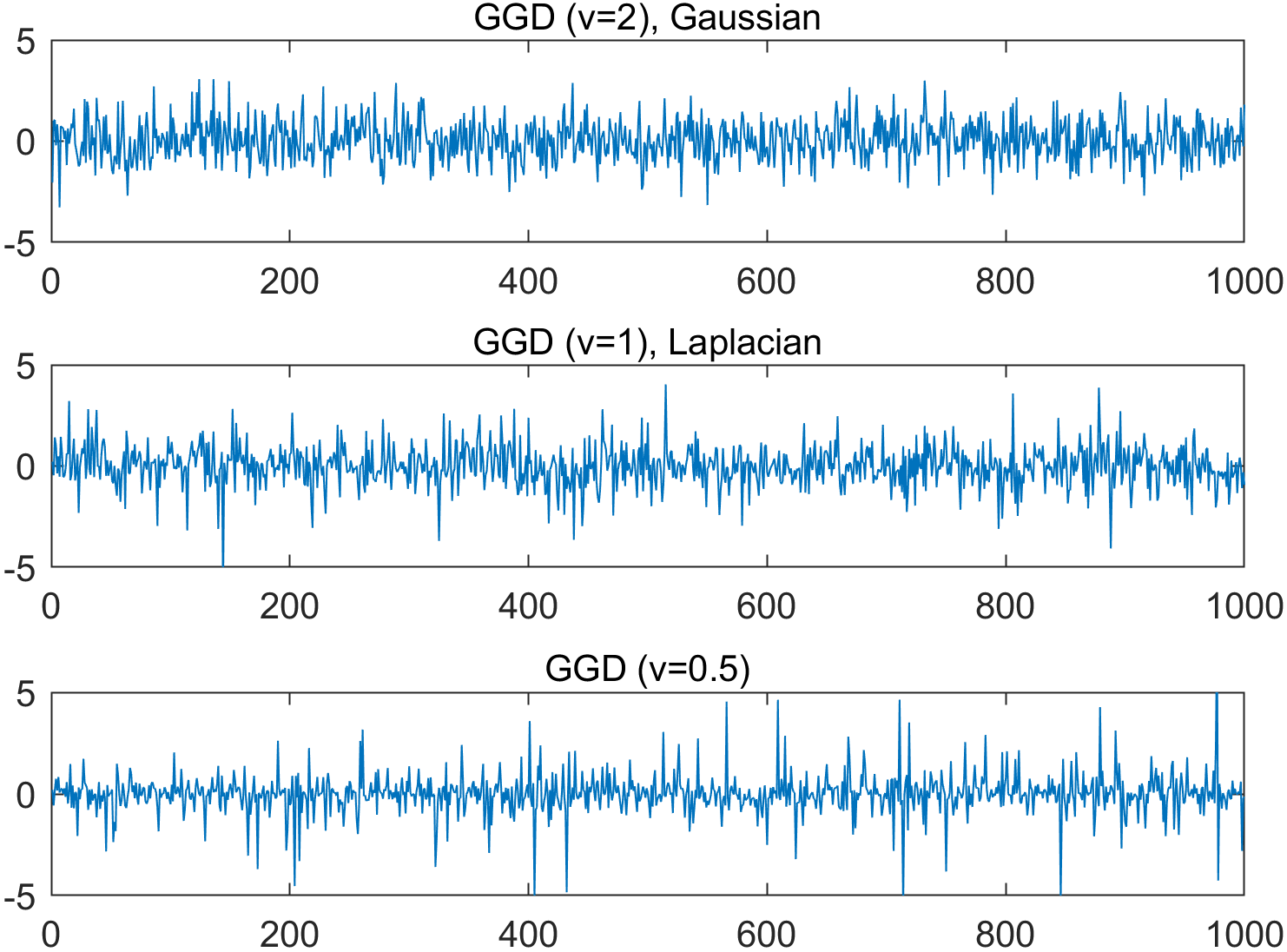}}
\caption{GGD noise with different shape parameter (with standard deviation of $\sigma=1$).}
\label{figure14}
\end{figure}

In the experiment, three noise cases with Gaussian and GGD ($v\in\{0.5,1\}$) inlier noise are considered. Figure~\ref{figure14} illustrates the three considered noises with standard deviation $\sigma=1$, including Gaussian, GGD with $v=0.5$ and GGD with $v=1$. For the GGD $p(x;v,\sigma)$, the standard deviation is given by
\[
\sigma_x=\sigma\sqrt{\frac{\Gamma(3/v)}{\Gamma(1/v)}}.
\]

We consider two noise strengths with $\sigma\in\{0.01,0.1\}$. Figure~\ref{figure12} shows the performance of the AM algorithm using the truncated LS and truncated $\ell_p$ loss in the linear regression experiment, whilst Figure~\ref{figure13} shows that in the rotation registration experiment. Four different values of $p$ are evaluated for the $\ell_p$ loss, with $p\in\{1,1.2,1.5,1.8\}$. It can be seen from Figure~\ref{figure12} and Figure~\ref{figure13} that in the case of GGD noise with $v=0.5$, the truncated $\ell_p$ loss with a relatively small value of $p$ (e.g. $p=1$) can yield significantly better accuracy than the truncated LS loss. Besides, in the condition of Laplacian noise, i.e. GGD noise with $v=1$, the truncated $\ell_p$ loss also yields distinctly better performance than the truncated LS loss in most cases.

\bibliographystyle{IEEEtran}

\end{document}